\documentclass[final]{cvpr}

\usepackage{times}
\usepackage{epsfig}
\usepackage{graphicx}
\usepackage{booktabs}
\usepackage{soul}
\usepackage{url}
\usepackage{adjustbox}
\usepackage[utf8]{inputenc}
\usepackage[small]{caption}
\usepackage{booktabs}
\usepackage{algorithm}
\usepackage[caption=false]{subfig}
\usepackage{algorithmic}
\usepackage{multirow}
\usepackage{amsmath,amsfonts,amssymb, amsthm}
\usepackage[hidelinks]{hyperref}
\usepackage{url}
\hypersetup{colorlinks,citecolor=blue}
\newcommand{\pnamefull}{{\tt Visual Question Answering and Reasoning}\xspace}
\newcommand{\pname}{{\tt ViQAR}}
\def\cvprPaperID{11} % *** Enter the CVPR Paper ID here
\def\confYear{CVPR 2021}
\newcommand{\beginsupplement}{%
        \setcounter{table}{0}
        \renewcommand{\thetable}{S\arabic{table}}%
        \setcounter{figure}{0}
        \renewcommand{\thefigure}{S\arabic{figure}}%
        \setcounter{section}{0}
        \renewcommand{\thesection}{S\arabic{section}}
     }

\begin{document}

%%%%%%%%% TITLE
\title{Beyond VQA: Generating Multi-word Answers and Rationales to Visual Questions}
% \author{{Radhika Dua, Sai Srinivas Kancheti, Vineeth N Balasubramanian}\\
% Indian Institute of Technology Hyderabad, India\\
% {\tt\small \{radhika,cs21resch01004,vineethnb\}@iith.ac.in}
% \and
% Sai Srinivas Kancheti\\
\author{{Radhika Dua\thanks{equal contribution}
\hspace*{5mm} Sai Srinivas Kancheti\footnotemark[1]\hspace*{5mm} Vineeth N Balasubramanian}\\
Indian Institute of Technology Hyderabad, India\\
{\tt\small \{radhika,cs21resch01004,vineethnb\}@iith.ac.in}
}
\maketitle
\begin{abstract}
Visual Question Answering is a multi-modal task that aims to measure high-level visual understanding. Contemporary VQA models are restrictive in the sense that answers are obtained via classification over a limited vocabulary (in the case of open-ended VQA), or via classification over a set of multiple-choice-type answers. In this work, we present a completely generative formulation where a multi-word answer is \textit{generated} for a visual query. To take this a step forward, we introduce a new task: \textbf{\pname}\ (Visual Question Answering and Reasoning), wherein a model must generate the complete answer and a rationale that seeks to justify the generated answer. We propose an end-to-end architecture to solve this task and describe how to evaluate it. We show that our model generates strong answers and rationales through qualitative and quantitative evaluation, as well as through a human Turing Test.

% \keywords{Visual Commonsense Reasoning \and Visual Question Answering \and Model with Generated Explanations}
\end{abstract}
% \footnote{$^*$ indicates equal contribution\\Corresponding author: Radhika Dua (\href{mailto:radhika@iith.ac.in}{radhika@iith.ac.in})}
%\vspace{-18pt}
\section{Introduction}
\label{sec:intro}
%\vspace{-7pt}

Visual Question Answering (VQA)~\cite{VQA} is a vision-language task that has seen a lot of attention in recent years. In general, the VQA task consists of either open-ended or multiple choice answers to a question about the image.
There are an increasing number of models that obtain the best possible performance on benchmark VQA datasets, which intend to measure visual understanding based on visual questions. However, answers in existing VQA datasets and models are largely one-word answers (average length is 1.1 words), which gives existing models the freedom to treat answer generation as a classification task. For the open-ended VQA task, the top-K answers are chosen, and models perform classification over this vocabulary. 
 
\begin{figure*}
\centering
\includegraphics[width =0.98\textwidth]{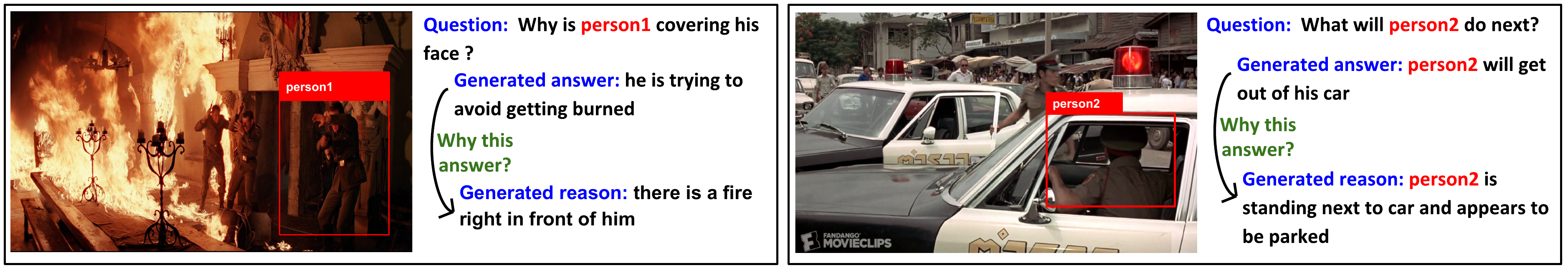}
% \vspace{-6pt}
\caption{Given an image and a question about the image, we \textbf{generate} a natural language answer and reason that explains why the answer was generated. The images shown above are examples of outputs that our proposed model generates. 
%To the best of our knowledge, this is the first such effort of automatically generating a multi-word answer and rationale to a visual question, instead of picking answers from a pre-defined list.
These examples also illustrate the kind of visual questions for which a single-word answer is insufficient. Contemporary VQA models handle even such kinds of questions only in a classification setting, which is limiting.}
\label{fig:VCR_generative}
% \vspace{-5pt}
\end{figure*}

However, many questions which require commonsense reasoning cannot be answered in a single word. A textual answer for a sufficiently complicated question may need to be a sentence. For example, a question of the type "What will happen...." usually cannot be answered completely using a single word. Figure~\ref{fig:VCR_generative} shows examples of such questions where multi-word answers are required (the answers and rationales in this figure are generated by our model in this work). Current VQA systems are not well-suited for questions of this type. To reduce this gap, more recently, the Visual Commonsense Reasoning (VCR) task~\cite{Zellers2018FromRT,Lu2019ViLBERTPT,Dua2019DROPAR,Zheng2019ReasoningVD,Talmor2018CommonsenseQAAQ} was proposed, which requires a greater level of visual understanding and an
% \begin{figure*}
% \centering
% %\vspace{-6pt}
% % %\vspace{-17pt}
% \includegraphics[width =0.95\textwidth,height=0.22\textwidth ]{figures/many_ans_5739_new.pdf}
% %\vspace{-4pt}
% \caption{An example from the VCR dataset \cite{Zellers2018FromRT} shows that there can be many correct multi-word answers to a question, which makes classification setting restrictive. The highlighted option is the correct option present in the VCR dataset, the rest are examples of plausible correct answers.}
% % %\vspace{-17pt}
% %\vspace{-10pt}
% \label{fig:VCR_example_manychoices}
% \end{figure*}
\begin{figure*}
\centering
\includegraphics[width =0.85\textwidth ]{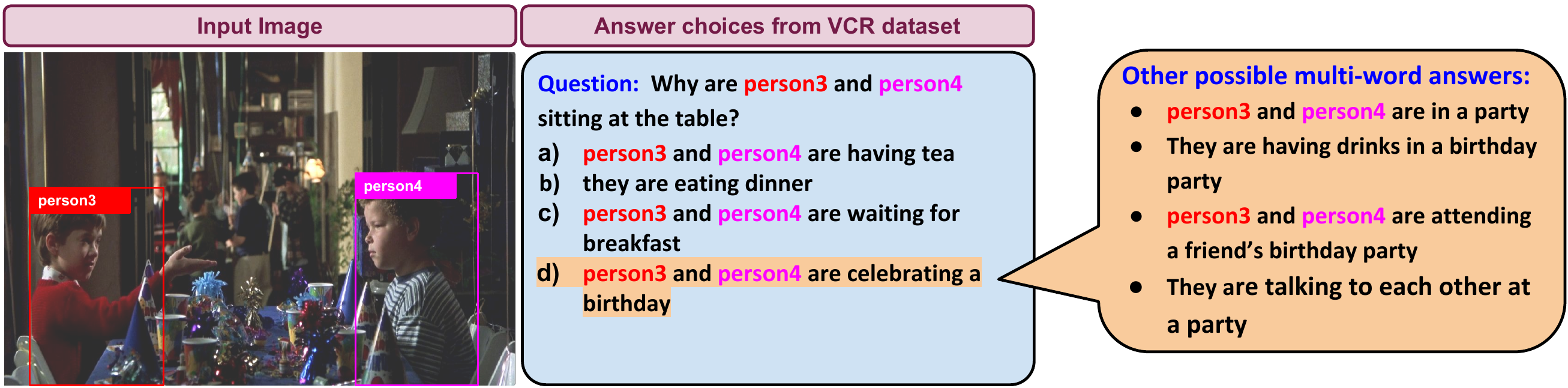}
% \vspace{-6pt}
\caption{An example from the VCR dataset shows that there can be many correct multi-word answers to a question that makes classification setting restrictive. The highlighted option is the correct option present in the VCR dataset, the rest are plausible correct answers.}
\label{fig:VCR_example_manychoices}
% \vspace{-6pt}
\end{figure*}
ability to reason about the world. More interestingly, the VCR dataset features multi-word answers, with an average answer length of $7.55$ words. However, VCR is still a classification task, where the correct answer is chosen from a set of four answers. Models which solve classification tasks simply need to pick an answer in the case of VQA, or an answer and a rationale for VCR. However, when multi-word answers are required for a visual question, options are not sufficient, since the same 'correct' answer can be paraphrased in a multitude of ways, each having the same semantic meaning but differing in grammar. Figure \ref{fig:VCR_example_manychoices} shows an image from the VCR dataset, where the first highlighted answer is the correct one among a set of four options provided in the dataset. The remaining three answers in the figure are included by us here (not in the dataset) as other plausible correct answers. Existing VQA models are fundamentally limited by picking a right option, rather to answer in a more natural manner. % Besides, classification models for multi-word answers may be prone to memorization or lack high-level reasoning.  
% \begin{figure*}
% \centering
% \subfloat{\includegraphics[width=0.49\linewidth,height=2.2cm]{figures/vcr_many_ans(5739).pdf}\label{fig:q1}}\hfill
% \subfloat{\includegraphics[width=0.49\linewidth,height=2.2cm]{figures/vcr_many_ans(3440).pdf}\label{fig:q1}}\\
% %\vspace{-9pt}
% \caption{Examples from the VCR dataset to demonstrate that there can be many more correct answers for a question, which makes the classification setting very restrictive.}
% \label{fig:VCR_example_manychoices}
% \end{figure*}
Moreover, since the number of possible `correct' options in multi-word answer settings can be large (as evidenced by Figure \ref{fig:VCR_example_manychoices}), we propose that for richer answers, one would need to move away from the traditional classification setting, and instead let our model \textit{generate the answer} to a given question. We hence propose a new task which takes a generative approach to multi-word VQA in this work. 

Humans when answering questions often use a rationale to justify the answer. In certain cases, humans answer directly from memory (perhaps through associations) and then provide a post-hoc rationale, which could help improve the answer too - thus suggesting an interplay between an answer and its rationale. Following this cue, we also propose to generate a rationale along with the answer which serves two purposes: (i) it helps justify the generated answer to end-users; and (ii) it helps generate a better answer.
Going beyond contemporary efforts in VQA, we hence propose, for the first time to the best of our knowledge, an approach that automatically generates both multi-word answers and an accompanying rationale, that also serves as a textual justification for the answer. We term this task  \pnamefull (\pname) and propose an end-to-end methodology to address this task.  This task is especially important in critical AI tasks such as VQA in the medical domain, where simply answering questions about the medical images is not sufficient. 
In addition to formalizing this new task, we provide a simple yet reasonably effective model consisting of four sequentially arranged recurrent networks to address this challenge. The model can be seen as having two parts: a generation module (GM), which comprises of the first two sequential recurrent networks, and a refinement module (RM), which comprises of the final two sequential recurrent networks. The GM first generates an answer, using which it generates a rationale that explains the answer. The RM generates a \textit{refined} answer based on the rationale generated by GM. The refined answer is further used to generate a refined rationale. Our overall model design is motivated by the way humans think about answers to questions, wherein the answer and rationale are often mutually dependent on each other. We seek to model this dependency by first generating an answer-rationale pair and then using them as priors to regenerate a refined answer and rationale. We train our model on the VCR dataset, which contains open-ended visual questions along with answers and rationales. Considering this is a generative task, we evaluate our methodology by comparing our generated answer/rationale with the ground truth answer/rationale on correctness and goodness of the generated content using generative language metrics, as well as by human Turing Tests. %Considering the difficulty of this task, we observe promising qualitative and quantitative results, as presented later in this paper.

%\vspace{-1mm}
Our main contributions in this work can be summarized as follows: (i) We propose a new task \pname\ that seeks to open up a new dimension of Visual Question Answering tasks, by moving to a completely generative paradigm; (ii) We propose a simple and effective model based on generation and iterative refinement for \pname\ (which could serve as a baseline to the community); (iii) Considering generative models in general can be difficult to evaluate, we provide a discussion on how to evaluate such models, as well as study a comprehensive list of evaluation metrics for this task; (iv) We conduct a suite of experiments which show promise of the proposed model for this task, and also perform ablation studies of various choices and components to study the effectiveness of the proposed methodology on \pname. %\redc{In summary the main focus of this work is to propose a new generative task that goes beyond traditional VQA, solve it with a simple yet effective baseline model, and provide a baseline for further efforts.} 
We believe that this work could lead to further efforts on common-sense answer and rationale generation in vision tasks in the near future. 
To the best of our knowledge, this is the first such effort of automatically generating a multi-word answer and rationale to a visual question.
\section{Related Work}\label{sec:related}
% In this section, we review earlier efforts from multiple perspectives that may be related to this work: Visual Question Answering, Visual Commonsense Reasoning and Image Captioning in general.
\paragraph{VQA and Image Captioning.}
% VQA and its variants have been the subject of much research work.
A lot of work in VQA is based on attention-based models that aim to 'look' at the relevant regions of the image in order to answer the question ~\cite{DBLP:journals/corr/AndersonHBTJGZ17,Lu2016HierarchicalQC,Yu2017MultilevelAN,Yi2018NeuralSymbolicVD}. Other recent work has focused on better multimodal fusion methods~\cite{Kim2018BilinearAN,DBLP:journals/corr/KimOKHZ16,Fukui2016MultimodalCB,Yu2017MultimodalFB}, the
incorporation of relations~\cite{NorcliffeBrown2018LearningCG,Li2019RelationAwareGA,Santoro2017ASN}, the use of multi-step reasoning~\cite{Cadne2019MURELMR},  and neural module
networks for compositional reasoning~\cite{Johnson2017InferringAE,Chen_2021_WACV,Hu2017LearningTR}. Visual Dialog~\cite{Das2018VisualD,Zheng2019ReasoningVD} extends VQA but requires an agent to hold a meaningful conversation with humans in natural language based on visual questions. Image captioning~\cite{Xu2015ShowAA,You2016ImageCW,Lu2016KnowingWT,DBLP:journals/corr/AndersonHBTJGZ17,Rennie2016SelfCriticalST} is a global description of an image and hence different from \pname\ which is concerned with answering a question about understanding of a local region in the image.

The efforts closest to ours are those that provide justifications along with answers~\cite{Li2018VQAEEE,Hendricks2016GeneratingVE,Li2018TellandAnswerTE,Park2018MultimodalEJ,Wu2019GeneratingQR,Park2018MultimodalEJ}, each of which however also answers a question as a classification task (and not in a generative manner) as described below. Li et al.~\cite{Li2018VQAEEE} created the VQA-E dataset that has an explanation along with the answer to the question. Wu et al.~\cite{Wu2019GeneratingQR} provide relevant captions to aid in solving VQA, which can be thought of as weak justifications. More recent efforts~\cite{Park2018MultimodalEJ,Patro2020RobustEF} attempt to provide visual and textual explanations to justify the predicted answers. Datasets have also been proposed for VQA in the recent past to test visual understanding \cite{Zhu2015Visual7WGQ,Goyal2016MakingTV,Johnson2016CLEVRAD}; for e.g., the Visual7W dataset~\cite{Zhu2015Visual7WGQ} contains a richer class of questions about an image with textual and visual answers. However, all these aforementioned efforts continue to focus on answering a question as a classification task (often in one word, such as Yes/No), followed by simple explanations. We however, in this work, focus on \textit{generating} multi-word answers with a corresponding multi-word rationale, which has not been done before. 
% However, the answers in these are open-ended(generally yes/no, numerical, single-word answers), and the problem persists since all the questions cannot be answered with one word and existing work (VQA-E and Multi-modal explanation) fails to provide  multi-word answers.
%\textbf{BERT: }Bidirectional Encoder Representations from Transformers~\cite{Devlin2019BERTPO} state-of-the-art results in a wide variety of NLP tasks, including Question Answering. BERT can be used to obtain sentence embeddings that are contextual. The vector representation of a token depends on the other words in the sequence, in both directions. We have used the same to obtain sentence level contextual embedding for give question and image caption. Contextual embeddings capture greater linguistic information than token level embeddings.
% 
\vspace{-6pt}
\paragraph{Visual Commonsense Reasoning (VCR).}
VCR \cite{Zellers2018FromRT} is a vision-language dataset, which involves choosing a correct answer (among four provided options) for a given question about the image, and then choosing a rationale to justify the answer. The task associated with the dataset aims to test for visual commonsense understanding and provides images, questions and answers of a higher complexity than other datasets such as CLEVR \cite{Johnson2016CLEVRAD}. The dataset has attracted various methods~\cite{Zellers2018FromRT,Lu2019ViLBERTPT,Dua2019DROPAR,Zheng2019ReasoningVD,Talmor2018CommonsenseQAAQ,Lin2019TABVCRTA,Brad2019SceneGC}, each of which however follow the dataset's task and treat this as a classification problem. None of these efforts attempt to answer and reason using generated sentences.%
%\vspace{-12pt}
%Such a task has an additional layer of complexity to it, since first and foremost, the generated text has to be structurally correct. Further, an image captioning model needs to generate grammatically correct captions, even if its visual understanding is lacking. %Most captioning architectures are attention based encoder-decoder architectures ~\cite{Xu2015ShowAA,You2016ImageCW,Lu2016KnowingWT,Anderson2017BottomUpAT, Rennie2016SelfCriticalST}.
  %\cite{Anderson2017BottomUpAT} introduce object-level image features as bottom-up-attention. ~\cite{Lu2016KnowingWT} proposes soft spatial attention, which is the top-down attention of choice in ~\cite{Anderson2017BottomUpAT}. 
% 

In contrast to all the aforementioned efforts, our work, \pname, focuses on automatic complete \textit{generation} of the answer, and of a rationale, given a visual query. This is a challenging task, since the generated answers must be correct (with respect to the question asked), be complete, be natural, and also be justified with a well-formed rationale.
% We now describe the task, and our methodology for addressing this task.
%\vspace{-11pt}
\section{ViQAR: Task description}
\label{sec_task_description}
%\vspace{-7pt}
%In this paper, we propose \textbf{\pname: \pnamefull}, a novel task for generating an answer and a reason given an image, a caption, and a question. 
Let $\mathcal{V}$ be a given vocabulary of size $|\mathcal{V}|$ and $\mathbf{A}~=~(a_1, \ldots ,  a_{l_a}) \in \mathcal{V}^{l_a}$, $\mathbf{R}~=~(r_1, \ldots, r_{l_r}) \in \mathcal{V}^{l_r}$ represent answer sequences of length $l_a$ and rationale sequences of length $l_r$ respectively. %, where $a_i$ and $r_i$ are discrete random variables taking values from $\mathcal{V}$. 
%Let $V = \{\bold{v}_1, \bold{v}_2, ... , \bold{v}_k\}$, $\bold{v}_i\in \mathbb{R}^D$ be the set of $k$ spatial image features,
Let $\mathbf{I} \in \mathbb{R}^D$ represent the image representation, and
$\mathbf{Q}\in\mathbb{R}^B$ be the feature representation of a given question. We also allow the use of an image caption, if available, in this framework given by a feature representation  $\mathbf{C}\in\mathbb{R}^B$. Our task is to compute a function $\mathcal{F}:\mathbb{R}^D\times\mathbb{R}^B\times\mathbb{R}^B\rightarrow\mathcal{V}^{l_a}\times\mathcal{V}^{l_r}$ that maps the input image, question and caption features to a large space of generated answers $\mathbf{A}$ and rationales $\mathbf{R}$, as given below:
% \begin{ceqn}
\begin{equation}
%\vspace{-3pt}
    \mathcal{F}(\mathbf{I},\mathbf{Q},\mathbf{C})=(\mathbf{A},\mathbf{R})
%%\vspace{-1pt}
\end{equation}
% \end{ceqn}
\noindent %where $\bold{I}$ represents the image features. 
Note that the formalization of this task is different from other tasks in this domain, such as Visual Question Answering \cite{VQA} and Visual Commonsense Reasoning \cite{Zellers2018FromRT}. The VQA task can be formulated as learning a function $\mathcal{G}:\mathbb{R}^D\times\mathbb{R}^B\rightarrow C$, where $C$ is a discrete, finite set of choices (classification setting). Similarly, the Visual Commonsense Reasoning task provided in \cite{Zellers2018FromRT} aims to learn a function $\mathcal{H}:\mathbb{R}^D\times\mathbb{R}^B\rightarrow C_1\times C_2$, where $C_1$ is the set of possible answers, and $C_2$ is the set of possible reasons. The generative task, proposed here in \pname, is harder to solve when compared to VQA and VCR. One can divide \pname\ into two sub-tasks:
%\vspace{-4pt}
\begin{itemize}
\item \textbf{Answer Generation.} Given an image, its caption, and a complex question about the image, a multi-word natural language answer is generated: \\
% \begin{center}
$(\mathbf{I}, \mathbf{Q}, \mathbf{C}) \rightarrow \mathbf{A}$ 
% \end{center}
\item \textbf{Rationale Generation.}
 Given an image, its caption, a complex question about the image, and an answer to the question, a rationale to justify the answer is generated: 
% \begin{center}
$(\mathbf{I}, \mathbf{Q}, \mathbf{C}, \mathbf{A}) \rightarrow  \mathbf{R}$
% \end{center}
\end{itemize}
%\vspace{-5pt}
We also study variants of the above sub-tasks (such as when captions are not available) in our experiments. Our experiments suggest that the availability of captions helps a model achieve better performance on our task. %The above formulation is a generic framework that can include captions(which we have found to help performance). We show results with and without them. 
We now present a methodology built using known basic components to study and show that the proposed, seemingly challenging, new task can be solved with existing architectures. In particular, our methodology is based on the understanding that the answer and rationale can help each other, and hence needs an iterative refinement procedure to handle such a multi-word multi-output task. We consider the simplicity of the proposed solution as an aspect of our solution by design, more than a limitation, and hope that the proposed architecture will serve as a baseline for future efforts on this task. 
%integrate them into our final architecture. We now present our \redc{simple yet effective} methodology for this task, which we hope will serve as a baseline for future efforts on this task. \redc{Our work shows that a relatively complex task like \pname can be solved using simple existing modules that do not require great computational overhead. This is an aspect of our solution rather than a limitation.}

%\vspace{-12pt}
\section{Proposed methodology}
\label{sec:Methodology}
% \vspace{-7pt}
% \begin{figure}
% \centering
% \includegraphics[width =\columnwidth]{figures/VCR_tsubscript.png}
% \caption{The decoder of our proposed architecture: Given an image and a question on the image, the model must generate an answer to the question and  a rationale to justify why the answer is correct.}
% \label{fig:proposed_architecture}
% \end{figure}
\begin{figure*}[hbt]
\centering
\includegraphics[width =0.9\textwidth]{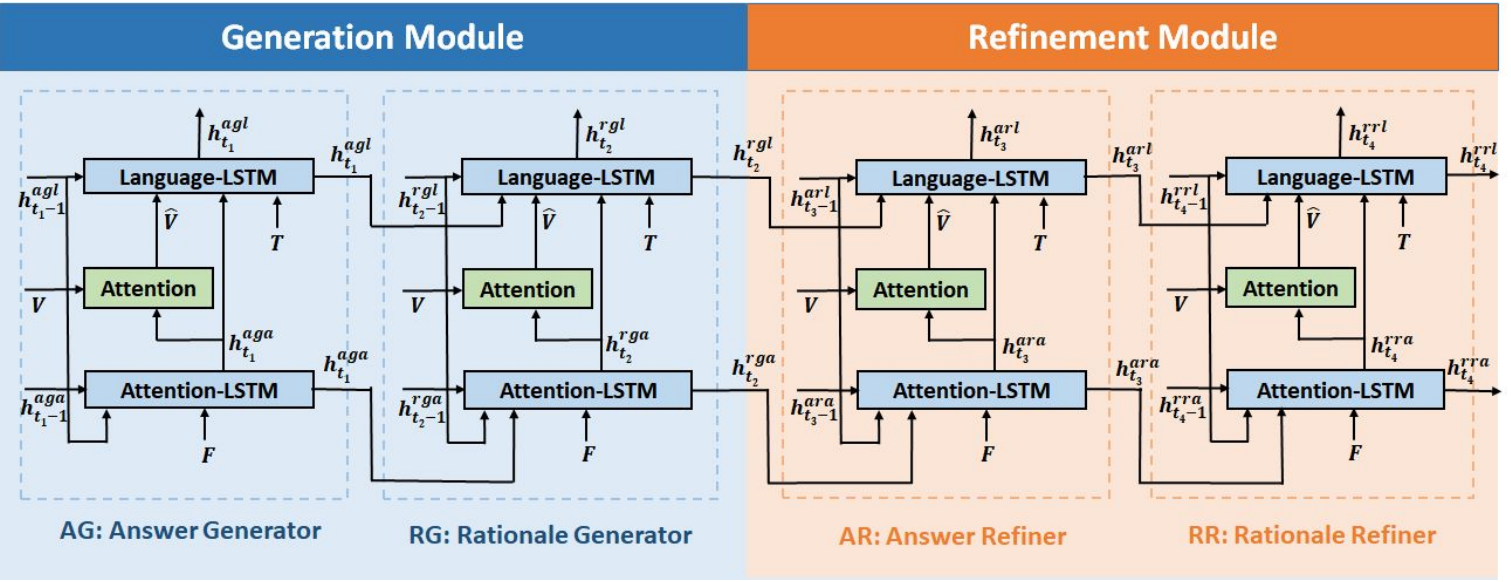}
\caption{ The decoder of our proposed architecture: For simplicity we only show the last time-step of each unrolled LSTM. Here $t_1=l_a$, $t_2=l_a+l_r$, $t_3=2l_a+l_r$ and $t_4=2l_a+2l_r$. Given an image and a question on the image, the model must generate an answer to the question and  a rationale to justify why the answer is correct.}
\label{fig:proposed_architecture}
\end{figure*}
We present an end-to-end, attention-based,  encoder-decoder architecture for answer and rationale generation which is based on an iterative refinement procedure. The refinement in our architecture is motivated by the observation that answers and rationales can influence one another mutually. Thus, knowing the answer helps in generation of a rationale, which in turn can help in the generation of a more refined answer. The encoder part of the architecture generates the features from the image, question and caption. These features are used by the decoder to generate the answer and rationale for a question.
%\vspace{-10pt}
\paragraph{Feature Extraction.}
% \label{sec:feature_extraction}
We use spatial image features as proposed in ~\cite{DBLP:journals/corr/AndersonHBTJGZ17}, which are termed bottom-up image features. We consider a fixed number of regions for each image, and extract a set of $k$ features, $V$, as defined below:
% \begin{ceqn}
\begin{align}\label{eqn_img_features} 
    V = \{\mathbf{v}_1,\mathbf{v}_2,\ldots,\mathbf{v}_k\} \quad \text{where} \quad \mathbf{v}_i\in \mathbb{R}^{D}.
\end{align}
% \end{ceqn}

We use BERT ~\cite{Devlin2019BERTPO} representations to obtain fixed-size embeddings for the question and caption, $\mathbf{Q}\in\mathbb{R}^B$ and $\mathbf{C}\in\mathbb{R}^B$ respectively. The question and caption are projected into a common feature space $\mathbf{T}\in\mathbb{R}^L$ given by:
% %\vspace{-3pt}
\begin{equation}
    \mathbf{T} = g(W_t^{T}({\rm tanh}(W_q^{T}\mathbf{Q})\oplus {\rm tanh}(W_c^{T}\mathbf{C}))),
\end{equation}
where $g$ is a non-linear function, $\oplus$ indicates concatenation and $W_t\in\mathbb{R}^{L\times L}$, $W_q\in\mathbb{R}^{B\times L}$ and $W_c\in\mathbb{R}^{B\times L}$ are learnable weight matrices of the layers (we use two linear layers in our implementation in this work).

Let the mean of the extracted spatial image features (as in Equation \ref{eqn_img_features}) be denoted by $\mathbf{\Bar{V}}\in\mathbb{R}^{D}$. These are concatenated with the projected question and caption features to obtain $\mathbf{F}$, which is the common input feature vector to all the LSTMs in our architecture:
%\vspace{-3pt}
\begin{equation}
    \mathbf{F} = \mathbf{\Bar{V}} \oplus \mathbf{T}
    \label{eqn:common_input}
%\vspace{-2pt}
\end{equation}
% \st{where $\mathbf{\Bar{V}}\in\mathbb{R}^{D}$ are the mean spatial image features.}
% We use $\mathbf{F}$ as 

%\vspace{-12pt}
\paragraph{Architecture.} Figure ~\ref{fig:proposed_architecture} shows our end-to-end architecture to address \pname. As stated earlier, our architecture has two modules: \textit{generation} ($GM$) and \textit{refinement} ($RM$). The $GM$ consists of two sequential, stacked LSTMs, henceforth referred to as answer generator ($AG$) and rationale generator ($RG$) respectively. The $RM$ seeks to refine the generated answer as well as rationale, and is an important part of the proposed solution as seen in our experimental results. It also consists of two sequential, stacked LSTMs, which we denote as answer refiner ($AR$) and rationale refiner ($RR$).

Each sub-module (presented inside dashed lines in the figure) is a complete LSTM. Given an image, question, and caption, the $AG$ sub-module unrolls for $l_a$ time steps to generate an answer. The hidden state of Language and Attention LSTMs after $l_a$ time steps is a representation of the generated answer. Using the representation of the generated answer from $AG$, $RG$ sub-module unrolls for $l_r$ time steps to generate a rationale and obtain its representation. Then the $AR$ sub-module uses the features from $RG$ to generate a refined answer. Lastly, the $RR$ sub-module uses the answer features from $AR$ to generate a refined rationale. Thus, a refined answer is generated after $l_a + l_r$ time steps and a refined rationale is generated after $l_a$ further time steps. The complete architecture runs in $2l_a + 2l_r$ time steps.

%\vspace{-10pt}
\paragraph{The LSTMs.} The two layers of each stacked LSTM \cite{Hochreiter1997LongSM} are referred to as the Attention-LSTM ($\mathcal{L}_a$) and Language-LSTM ($\mathcal{L}_l$) respectively. %, following the naming convention in ~\cite{DBLP:journals/corr/AndersonHBTJGZ17}. 
We denote $h_t^{a}$ and $x_t^{a}$ as the hidden state and input of the Attention-LSTM at time step $t$ respectively. Analogously, $h_t^{l}$ and $x_t^{l}$ denote the hidden state and input of the Language-LSTM at time $t$.
% The input to the encoder are the image, its caption and a question about the image. 
Since the four LSTMs are identical in operation, we describe the attention and sequence generation modules of one of the sequential LSTMs below in detail.

%\vspace{-10pt}
\paragraph{Spatial visual attention.}
We use a soft, spatial-attention model, similar to  ~\cite{DBLP:journals/corr/AndersonHBTJGZ17}
and ~\cite{Lu2016KnowingWT}, to compute attended image features $\mathbf{\hat{V}}$. Given the combined input features $\mathbf{F}$ and previous hidden states $h_{t-1}^{a}$, $h_{t-1}^{l}$, the current hidden state of the Attention-LSTM is given by:
%\vspace{-3pt}
\begin{align}
        x_t^a &\equiv h^p\oplus h_{t-1}^l\oplus \mathbf{F}\oplus \pi_t, \nonumber\\
        h_t^a &= \mathcal{L}_a(x_t^a,h_{t-1}^a),
\end{align}
where $\pi_t=W_e^T\mathbf{1}_t$ is the embedding of the input word, $W_e\in\mathbb{R}^{|\mathcal{V}|\times E}$ is the weight of the embedding layer, and $\mathbf{1}_t$ is the one-hot representation of the input at time $t$. $h^p$ is the hidden representation of the previous LSTM (answer or rationale, depending on the current LSTM).

The hidden state $h_t^a$ and visual features $V$ are used by the attention module (implemented as a two-layered MLP in this work) to compute the normalized set of attention weights $\boldsymbol{\alpha}_t = \{\alpha_{1t},\ldots,\alpha_{kt}\}$ (where $\alpha_{it}$ is the normalized weight of image feature $\mathbf{v}_i$) as below:
% %\vspace{-6pt}
\begin{align}
    y_{i,t} &= W_{ay}^{T}(\tanh(W_{av}^{T}\mathbf{v}_{i} + W_{ah}^{T}h_{t}^a)), \nonumber\\
    \boldsymbol{\alpha}_t &= \text{softmax}(y_{1t}\ldots,y_{kt}).
\end{align}

In the above equations, $W_{ay}\in\mathbb{R}^{A\times 1}$, $W_{av}\in\mathbb{R}^{D\times A}$ and $W_{ah}\in\mathbb{R}^{H\times A}$ are weights learned by the attention MLP, $H$ is the hidden size of the LSTM and $A$ is the hidden size of the attention MLP. The attended image feature vector $\mathbf{\hat{V}}_t=\sum_{i=1}^{k}\boldsymbol{\alpha}_{it}\mathbf{v}_{i}$ is the weighted sum of all visual features.

%\vspace{-13pt}
\paragraph{Sequence generation.}
The attended image features $\mathbf{\hat{V}}_t$, together with $\mathbf{T}$ and $h_t^a$, are inputs to the language-LSTM at time $t$. We then have:
%\vspace{-3pt}
\begin{align}
        x_t^l &\equiv h^p\oplus \mathbf{\hat{V}}_t\oplus h_t^a\oplus \mathbf{T} \nonumber\\
        h_t^l &= \mathcal{L}_l(x_t^l,h_{t-1}^l) \nonumber\\
        y_t &= W_{lh}^Th_t^l + b_{lh} \nonumber\\
        p_t &= {\rm softmax}(y_t)
\end{align}
where $h^p$ is the hidden state of the previous LSTM, $h_t^l$ is the output of the Language-LSTM, $p_t$ is the conditional probability over words in $\mathcal{V}$ at time $t$. The word at time step $t$ is generated by a single-layered MLP with learnable parameters: $W_{lh}\in\mathbb{R}^{H\times |\mathcal{V}|}$, $b_{lh}\in\mathbb{R}^{|\mathcal{V}|\times 1}$.
The attention MLP parameters $W_{ay}$, $W_{av}$ and $W_{ah}$, and embedding layer's parameters $W_e$ are shared across all four LSTMs. 
% (We reiterate that although the architecture is based on well-known components, the aforementioned design decisions were obtained after significant study.)

\begin{figure}[H]
    \centering
    \includegraphics[width=0.88\columnwidth]{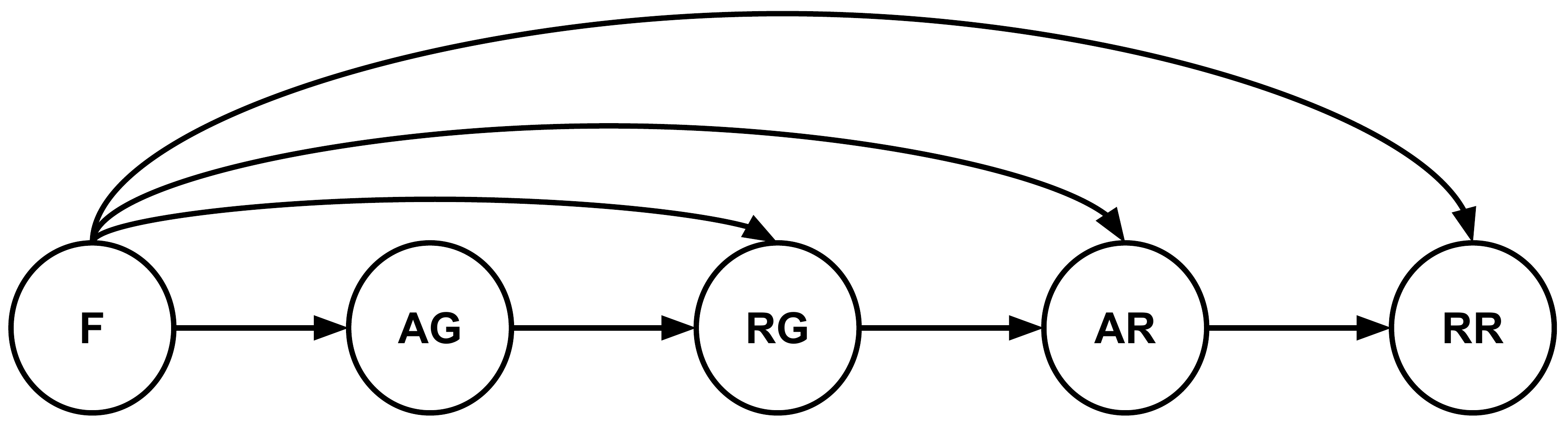}
    \caption{High-level illustration of our proposed Generation-Refinement model}
     \label{fig:proposed_architecture_alternate}
\end{figure}

\paragraph{Loss Function.} For a better understanding of our approach, Figure~\ref{fig:proposed_architecture_alternate} presents a high-level illustration of our proposed generation-refinement model.

Let $A_1 = (a_{11}, a_{12}, ..., a_{1l_a})$, $R_1 = (r_{11}, r_{12}, ..., r_{1l_r})$, $A_2 = (a_{21}, a_{22}, ..., a_{2l_a})$ and $R_2=(r_{21}, r_{22}, ..., r_{2l_r})$ be the generated answer, generated rationale, refined answer and refined rationale sequences respectively, where $a_{ij}$ and ${r_{ij}}$ are discrete random variables taking values from the common vocabulary $\mathcal{V}$. Given the common input $F$, our objective is to maximize the likelihood $P(A_1,R_1,A_2,R_2 | F)$ given by:
% \begin{ceqn}
\begin{align}
P(A_1, R_1, A_2, R_2 | F)&=P(A_1, R1 | F) P(A_2, R_2 | F, A_1, R_1) \nonumber\\
&=P(A_1|F)P(R_1|F,A_1)\nonumber\\& {}P(A_2|F,A_1,R_1)P(R_2 | F, A_1, R_1, A_2)\label{eqn:14}
% P(A_1, R_1, A_2, R_2 | F)&=P(A_1, R1 | F) P(A_2, R_2 | F, A_1, R_1) \nonumber\\
% &=P(A_1|F)P(R_1|F,A_1)P(A_2|F,A_1,R_1)P(R_2 | F, A_1, R_1, A_2)\label{eqn:14}
\vspace{-5pt}
\end{align}
% \end{ceqn}
In our model design, each term in the RHS of Eqn~\ref{eqn:14} is computed by a distinct LSTM. Hence, minimizing the sum of losses of the four LSTMs becomes equivalent to maximizing the joint likelihood. Our overall loss is the sum of four cross-entropy losses, one for each LSTM, as given below:
% \begin{ceqn}
\begin{equation}
% \begin{align}
\small
    \mathcal{L} = - \bigg( \sum_{t=1}^{l_a} \log p^{\theta_1}_t + \sum_{t=1}^{l_r} \log p^{\theta_2}_t + \sum_{t=1}^{l_a} \log p^{\theta_3}_t + \sum_{t=1}^{l_r} \log p^{\theta_4}_t\bigg)
% \end{align}
\end{equation}
% \end{ceqn}
\noindent where $\theta_i$ represents the $i^{th}$ sub-module LSTM, $p_t$ is the conditional probability of the $t^{th}$ word in the input sequence as calculated by the corresponding LSTM, $l_a$ indicates the ground-truth answer length, and $l_r$ the ground truth rationale length. Other loss formulations, such as a weighted average of the cross entropy terms did not perform better than a simple sum. We tried weights from ${0.0,0.25,0.5,0.75,1.0}$ for the loss terms.

\section{Experiments and results}
In this section, we describe the dataset used for this
work, implementation details of out model, and present the results of the proposed method and its variants.
\begin{table*}[hbt]
% \small
\begin{minipage}[t]{0.61\linewidth}
%   \caption{Quantitative evaluation on VCR dataset; CS = cosine similarity; we compare our model variants (last three columns) against defined baselines.
  \caption{Quantitative evaluation on VCR dataset; we compare against a basic two-stage LSTM model
and a VQA model as baselines; remaining columns are proposed model variants.[CS = cosine similarity]}
%   \vspace{-6pt}
\label{tab:vcr_gen_best}
\small
\resizebox{\linewidth}{!}{%
\begin{tabular}{lccccc} 
\toprule
 \textbf{Metrics}     & 
%  \textbf{VQA-Baseline}  &
 \begin{tabular}[c]{@{}c@{}}\textbf{VQA-Baseline}\\\textbf{}\end{tabular} &
 \begin{tabular}[c]{@{}c@{}}\textbf{Baseline}\\\textbf{}\end{tabular} &
 \begin{tabular}[c]{@{}c@{}}\textbf{Q+I+C}\\(Ours)\end{tabular} &
 \begin{tabular}[c]{@{}c@{}}\textbf{Q+I}\\(Ours)\end{tabular} &
 \begin{tabular}[c]{@{}c@{}}\textbf{Q+C}\\(Ours)\end{tabular} \\
%  \textbf{Baseline}  & 
%  \textbf{Q+I+C}  &
%  \textbf{Q+I}    & 
%  \textbf{Q+C}   \\ 
\midrule
Univ Sent Encoder CS  & 0.419                  & 0.410              & \textbf{0.455}  & 0.454           & 0.440          \\
Infersent CS          & 0.370                  & 0.400              & 0.438           & \textbf{0.442}  & 0.426          \\
Embedding Avg CS      & 0.838                  & 0.840              & 0.846           & \textbf{0.853}  & 0.845          \\
Vector Extrema CS     & 0.474                  & 0.444              & \textbf{0.493}  & 0.483           & 0.475          \\
Greedy Matching Score & 0.662                  & 0.633              & \textbf{0.672}  & 0.661           & 0.657          \\
METEOR                & 0.107                  & 0.095              & \textbf{0.116}  & 0.104           & 0.103          \\
Skipthought CS        & 0.430                  & 0.359              & \textbf{0.436}  & 0.387           & 0.385          \\
RougeL                & 0.259                  & 0.206              & \textbf{0.262}  & 0.232           & 0.236          \\
CIDEr                 & 0.364                  & 0.158              & \textbf{0.455}  & 0.310           & 0.298          \\
F-BERTScore           & 0.877                  & 0.860              & \textbf{0.879}  & 0.867           & 0.868          \\
\bottomrule
\end{tabular}}
  \end{minipage}\hfill
  \begin{minipage}[t]{0.36\linewidth}
\caption{Comparison of proposed Generation-Refinement architecture with variations in number of refinement modules. [CS: cosine similarity]}
% \vspace{-6pt}
\label{tab:vcr_gen_refinement}
\small
\resizebox{\linewidth}{!}{%
\begin{tabular}{lccc} 
\toprule
\multicolumn{1}{c}{ \textbf{Metrics} } & \multicolumn{3}{c}{\textbf{\#Refine Modules}}  \\ 
\cmidrule{2-4}
\multicolumn{1}{c}{}                   & \textbf{0}  & \textbf{1}      & \textbf{2}     \\ 
\midrule
Univ Sent Encoder CS                     & 0.453       & \textbf{0.455}  & 0.430          \\
Infersent CS                            & 0.434       & \textbf{0.438}  & 0.421          \\
Embedding Avg CS        & 0.850       & 0.846           & 0.840          \\
Vector Extrema CS       & 0.482       & \textbf{0.493}  & 0.462          \\
Greedy Matching Score                  & 0.659       & \textbf{0.672}  & 0.639          \\
METEOR                                 & 0.101       & \textbf{0.116}  & 0.090          \\
Skipthought CS          & 0.384       & \textbf{0.436}  & 0.375          \\
RougeL                                 & 0.234       & \textbf{0.262}  & 0.198          \\
CIDEr                                  & 0.314       & \textbf{0.455}  & 0.197          \\
F-BertScore                            & 0.868       & \textbf{0.879}  & 0.861          \\
\bottomrule
\end{tabular}
}
  \end{minipage}
%\vspace{-0.23in}
% \vspace{-5pt}
\end{table*}

\vspace{4mm}
\begin{table*}[hbt]
\centering
\caption{Results of the Turing test on VCR and Visual7W dataset performed with 30 people who had to rate samples consisting of a question and its corresponding answer and rationales on five criteria. For each criterion, a rating of 1 to 5 was given. The table gives the mean score and standard deviation for each criterion for the generated and ground truth samples.}
\label{tab:turing_test} 
% \vspace{-5pt}
\resizebox{\linewidth}{!}{%
\begin{tabular}{lcccc} 
\toprule
\multicolumn{1}{c}{\multirow{2}{*}{\begin{tabular}[c]{@{}c@{}} \textbf{}\\\textbf{Criteria}\end{tabular}}} & \multicolumn{2}{c}{\textbf{VCR}}                                                                               & \multicolumn{2}{c}{\textbf{Visual7W}}                                                                                                                             \\ 
\cmidrule{2-5}
\multicolumn{1}{c}{}                                                                                       & \begin{tabular}[c]{@{}c@{}}\textbf{Generated}\\\end{tabular} & \begin{tabular}[c]{@{}c@{}}\textbf{Ground-truth}\\\end{tabular} & \begin{tabular}[c]{@{}c@{}}\textbf{Generated}\\\end{tabular} & \begin{tabular}[c]{@{}c@{}}\textbf{Ground-truth}\\\end{tabular}  \\ 
\midrule
How well-formed and grammatically correct is the answer?                                                   & 4.15±1.05                   & 4.40±0.87                                                                        & 3.98±1.08                                                                     & --                                                                                 \\
How well-formed and grammatically correct is the rationale?                                                & 3.53±1.26                   & 4.26±0.92                                                                        & 3.80±1.04                                                                     & --                                                                                 \\
How relevant is the answer to the image-question pair?                                                     & 3.60±1.32                   & 4.08±1.03                                                                        & 4.11±1.17                                                                     & --                                                                                 \\
How well does the rationale explain the answer with respect to the image-question pair?                    & 3.04±1.36                   & 4.05±1.10                                                                        & 3.83±1.23                                                                     & --                                                                                 \\
Irrespective of the image-question pair, how well does the rationale explain the answer ?                  & 3.46±1.35                   & 4.13±1.09                                                                        & 3.83±1.28                                                                     & --                                                                                 \\
\bottomrule
\end{tabular}
}
\end{table*}
\label{sec:results}
\subsection{Experimental setup}
\paragraph{{Dataset.}}
Considering there has been no dataset explicitly built for this new task, we study the performance of the proposed method on the recently introduced VCR \cite{Zellers2018FromRT} dataset, which has all components needed for our approach. We train our proposed architecture on VCR, which contains ground truth answers and ground truth rationales against which we compare our generated answers and rationales.

\vspace{2mm}
\begin{table}[H]
\begin{center}
	\centering
		\caption{Statistical comparison of VCR with VQA-E, and VQA-X datasets. VCR dataset is highly complex as it is made up of complex subjective questions.}
	\footnotesize
	\begin{tabular}{lllll}
		\toprule
        \multirow{2}{*}\textbf{Dataset} & \textbf{Avg. A}  &\textbf{Avg. Q} &\textbf{Avg. R}  & \textbf{Complexity}  \\
         &\textbf{length} & \textbf{length} & \textbf{length} &{}\\
         \midrule
		 VCR & 7.55 &6.61 & 16.2 & High          \\
		 VQA-E  & 1.11 & 6.1 & 11.1 & Low       \\
		 VQA-X & 1.12 & 6.13 & 8.56 & Low       \\
		 \bottomrule
	\end{tabular}
	\label{tab:dataset_comparison}
\end{center}
\end{table}

VQA-E \cite{Li2018VQAEEE} and VQA-X \cite{Park2018MultimodalEJ} are competing datasets that contains explanations along with question-answer pairs. Table~\ref{tab:dataset_comparison} shows the high-level analysis of the three datasets. Since VQA-E and VQA-X are derived from VQA-2~\cite{Goyal2016MakingTV}, many of the questions can be answered in one word (a yes/no answer or a number). In contrast, VCR asks open-ended questions and has longer answers. Since our task aims to generate rich answers, the VCR dataset provides a richer context for this work.
CLEVR \cite{Johnson2016CLEVRAD} is another VQA dataset that measures the logical reasoning capabilities by asking the question that can be answered when a certain sequential reasoning is followed. This dataset however does not contain reasons/rationales on which we can train. Also, we do not perform a direct evaluation on CLEVR because our model is trained on real-world natural images while CLEVR is a synthetic shapes dataset.
In order to study our method further, we also study the transfer of our learned model to another challenging dataset, Visual7W \cite{Zhu2015Visual7WGQ}, by generating an answer/rationale pair for visual questions in Visual7W (please see Section \ref{sec:ablation} more more details).

\paragraph{{Implementation Details.}} We use spatial image features generated from~\cite{DBLP:journals/corr/AndersonHBTJGZ17} as our image input. Fixed-size BERT representations of questions and captions are used. Hidden size of all LSTMs is set to 1024 and hidden size of the attention MLP is set to 512. We trained using the ADAM optimizer with a decaying learning rate starting from $4e^{-4}$, using a batch size of 64. Dropout is used as a regularizer. 

%\vspace{-12pt}
\paragraph{{Evaluation metrics.}} We use multiple objective evaluation metrics to evaluate the goodness of answers and rationales generated by our model. 
%\noindent \textit{Automatic Evaluation Metrics}:
Since our task is generative, evaluation is done by comparing our generated sentences with ground-truth sentences to assess their semantic correctness as well as structural soundness. To this end, we use  a combination of multiple existing evaluation metrics. Word overlap-based metrics such as METEOR \cite{Lavie:2007:MAM:1626355.1626389},  CIDEr \cite{Vedantam2014CIDErCI} and ROUGE \cite{lin:2004:ACLsummarization} quantify the structural closeness of the generated sentences to the ground-truth. While such metrics give a sense of the structural correctness of the generated sentences, they may be insufficient for evaluating generation tasks, since there could be many valid generations which are correct, but not share the same words as a single ground truth answer. Hence, in order to measure how close the generation is to the ground-truth in meaning, we additionally use embedding-based metrics (which calculate the cosine similarity between sentence embeddings for generated and ground-truth sentences) including SkipThought cosine similarity~\cite{DBLP:journals/corr/KirosZSZTUF15}, Vector Extrema cosine similarity \cite{forgues2014bootstrapping}, Universal sentence encoder~\cite{DBLP:journals/corr/abs-1803-11175}, Infersent~\cite{DBLP:journals/corr/ConneauKSBB17} and BERTScore~\cite{Zhang2019BERTScoreET}. We use a comprehensive suite of all the aforementioned metrics to study the performance of our model. We further provide details of performance of our model on the VCR classification task in the supplementary.
\subsection{Performance evaluation of \pname}

\paragraph{Quantitative results.}
Quantitative results on the suite of evaluation metrics stated earlier are shown in Table \ref{tab:vcr_gen_best}. Since this is a new task, there are no known methods to compare against. We compare our model against a baseline (called  \textit{Baseline} in Table \ref{tab:vcr_gen_best}) composed of two separate two-stage LSTMs, one for answer and one for the rationale, and a VQA-based method ~\cite{DBLP:journals/corr/AndersonHBTJGZ17} that extracts multi-modal features to generate answers and rationales parallelly in an end-to-end manner (called \textit{VQA-Baseline} in Table \ref{tab:vcr_gen_best}). (Comparison with other standard VQA models is not relevant in this setting, since we perform a generative task, unlike existing VQA models.) We show results on three variants of our proposed Generation-Refinement model: Q+I+C (when question, image and caption are given as inputs), Q+I (question and image alone as inputs), and Q+C (question and caption alone as inputs). Evidently, our Q+I+C performed the most consistently across all the evaluation metrics. Importantly, our model outperforms both baselines, including the VQA-based one, on every single evaluation metric, showing the utility of the proposed architecture.

\begin{figure}[hbt]
\centering
\subfloat{\includegraphics[width=\columnwidth]{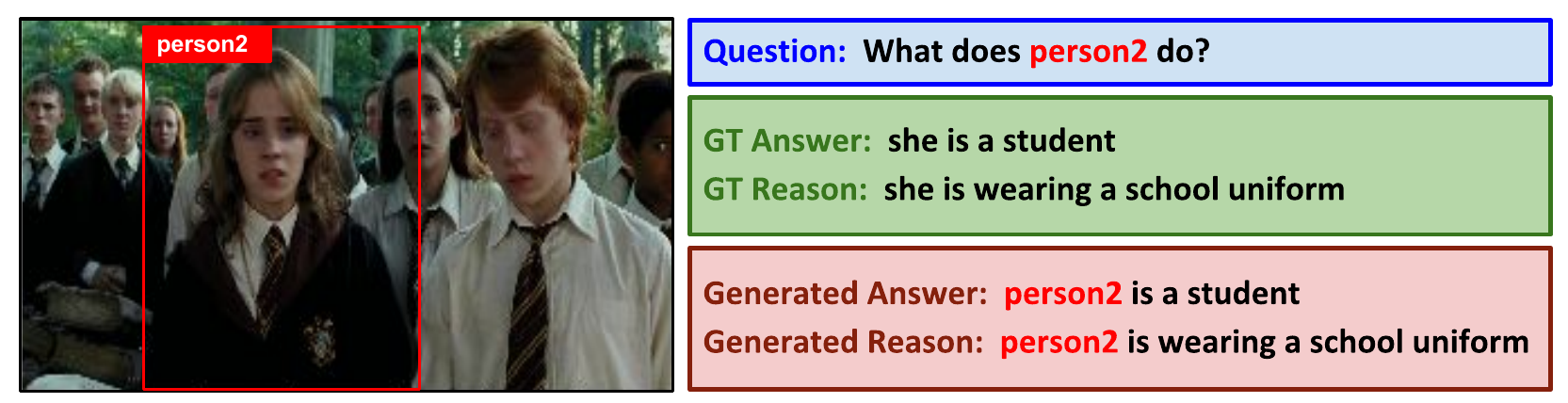}}
\vfill
\vspace{-6pt}
\subfloat{\includegraphics[width=\columnwidth]{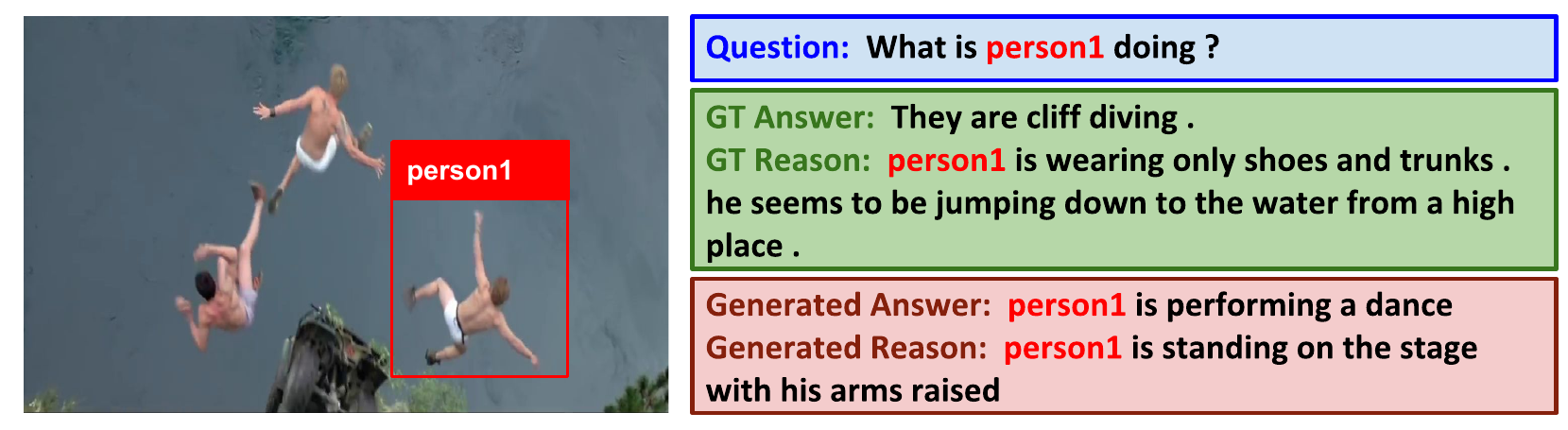}}
\caption{\textit{(Best viewed in color)} \textbf{Top:} Example output from our proposed Generation-Refinement architecture. \textbf{Bottom:} A challenging input for which our model fails.}
\label{fig:qualitative_results}
\end{figure}

\paragraph{Qualitative results.}
Figure~\ref{fig:qualitative_results} (Top) shows an example where the proposed model (Q+I+C setting) generates a meaningful answer with a supporting rationale.
%\bluec{Qualitative results indicate that our model is capable of generating answer-rationale pairs to complex subjective questions of the type: Explanation (why, how come), Activity (doing, looking, event), Temporal (happened, before, after, etc), Mental (feeling, thinking, love, upset), Scene (where, time) and Hypothetical sentences (if, would, could).}
Given the question, "What does person2 do?", the generated rationale: "person2 is wearing a school uniform" actively supports the generated answer: "person2 is a student", justifying the choice of a two-stage generation-refinement architecture.

% \bluec{Given the question, "What does person2 do?", the generated rationale: "person2 is wearing a school uniform" shows that the model generates the answer: "person2 is a student" because it can see that the person is wearing a school uniform which is usually worn by a student.}
For completeness of understanding, we present an example, Figure~\ref{fig:qualitative_results} (Bottom), on which our model fails to generate the semantically correct answer. Even on this result, we observe the generated answer and rationale are grammatically correct and complete (where rationale supports the answer). Improving the semantic correctness of the generations will be an important direction of future work. % Improving the semantic correctness of the generations will be an important direction of future work. %However, we observe that in these examples, the generated rationale is capable of justifying the incorrect answer, showing that a rationale is simply not memorized. 
Qualitative results indicate that our model is capable of generating answer-rationale pairs to complex subjective questions starting with 'Why', 'What', 'How', etc. More qualitative results
are presented in the supplementary owing to space constraints. 
\vspace{-2mm}
\paragraph{{Human Turing test.}}
In addition to the study of the objective evaluation metrics, we also performed a human Turing test on the generated answers and rationales. 30 human evaluators were presented each with 50 randomly sampled image-question pairs, each containing an answer to the question and its rationale. The test aims to measure how humans score the generated sentences w.r.t. ground truth sentences. Sixteen of the fifty questions had ground truth answers and rationales, while the rest were generated by our proposed model. For each sample, the evaluators had to give a rating of 1 to 5 for five different criteria, with 1 being very poor and 5 being very good. The results are presented in Table \ref{tab:turing_test}. Despite the higher number of generated answer-rationales judged by human users, the answers and rationales produced by our method were deemed to be fairly correct grammatically. The evaluators also agreed that our answers were relevant to the question and the generated rationales are acceptably relevant to the generated answer.
\subsection{Further analysis of \pname}
\label{sec:ablation}
%\vspace{-7pt}
%We study the proposed model under different settings to understand its efficacy, and present broader thoughts on automatic evaluation for such tasks.
%\vspace{-10pt} 
\paragraph{Ablation studies on refinement module.}
We evaluate the performance of variations of our proposed generation-refinement architecture $M$: (i) $M - RM$:  where the refinement module is removed; (ii) $M + RM$: where a second refinement module is added, i.e. the model has one generation and two refinement modules (to see if further refinement of answer and rationale helps). Table \ref{tab:vcr_gen_refinement} shows the quantitative results.
% \begin{table}[H]
% %\vspace{-6pt}
% \begin{center}
% 	\centering
% 	\footnotesize
% 		\begin{tabular}{|l|l|r|r|r|}
% 				\hline
% 		    \multirow{0}{*}{\textbf{Metrics}} &
% 		 	\multicolumn{3}{c|}{\textbf{\#Refine Modules}}\\
% % 			{\textbf{Metrics}} 
% 		    &\textbf{0} & \textbf{1}   & \textbf{2}     \\ \hline \hline
% 			Univ Sent Encoder
% 		    & 0.453 & \textbf{0.455} & 0.430	\\ \hline
% 			Infersent
% 			 & 0.434 & \textbf{0.438} & 0.421 	\\ \hline
% 			Embedding Avg Cosine similarity
% 			& 0.85 & 0.846 & 0.840	\\ \hline
% 			Vector Extrema Cosine Similarity
% 		    &  0.482 & \textbf{0.493} & 0.462 	\\ \hline
% 		    Greedy Matching Score
% 		    &  0.659 & \textbf{0.672} & 0.639 	\\ \hline
% 			METEOR
% 		    &  0.101 & \textbf{0.116} & 0.090   \\ 
% 		    \hline
% 		    Skipthought Cosine Similarity
% 			& 0.384 & \textbf{0.436} & 0.375 	\\ \hline
% 			RougeL
% 			&  0.234 & \textbf{0.262}&  0.198 	\\ \hline
% 			CIDEr
% 			&  0.314 & \textbf{0.455} & 0.197 	\\ \hline
% % 			P-BertScore
% % 			&  0.869 & \textbf{0.875} & 0.864 	\\ \hline
% % 			R-BertScore
% % 			&  0.868 & \textbf{0.885} & 0.857 	\\ \hline
% 			F-BertScore
% 			&  0.868 & \textbf{0.879} & 0.861 	\\ \hline
% 		\end{tabular}
% 	\caption{Comparison of proposed Generation-Refinement Architecture with variations in num of Refinement modules}
% 	\label{tab:vcr_gen_refinement}
% 	\end{center}
% 	%\vspace{-16pt}
% \end{table}
We observe that our proposed model, which has one refinement module has the best results. Adding additional refinement modules causes the performance to go down. 
% 
% \begin{wraptable}[18]{r}{0.6\textwidth} 
%\begin{table}[h]
% \begin{center}
%     %\vspace{-10pt}
%     \footnotesize
% 	\centering
% 		\begin{tabular}{|l|l|r|r|r|}
% 				\hline
% 		    \multirowcell{}{\textbf{Metrics}} &
% 		 	\multicolumn{3}{c|}{\textbf{\#Ref Modules}}\\
% % 			{\textbf{Metrics}} 
% 		    &\textbf{0} & \textbf{1}   & \textbf{2}     \\ \hline \hline
% 			Univ Sent Encoder
% 		    & 0.453 & \textbf{0.455} & 0.430	\\ \hline
% 			Infersent
% 			 & 0.434 & \textbf{0.438} & 0.421 	\\ \hline
% 			Embedding Avg Cosine similarity
% 			& 0.85 & 0.846 & 0.840	\\ \hline
% 			Vector Extrema Cosine Similarity
% 		    &  0.482 & \textbf{0.493} & 0.462 	\\ \hline
% 		    Greedy Matching Score
% 		    &  0.659 & \textbf{0.672} & 0.639 	\\ \hline
% 			METEOR
% 		    &  0.101 & \textbf{0.116} & 0.090   \\ 
% 		    \hline
% 		    Skipthought Cosine Similarity
% 			& 0.384 & \textbf{0.436} & 0.375 	\\ \hline
% 			RougeL
% 			&  0.234 & \textbf{0.262}&  0.198 	\\ \hline
% 			CIDEr
% 			&  0.314 & \textbf{0.455} & 0.197 	\\ \hline
% % 			P-BertScore
% % 			&  0.869 & \textbf{0.875} & 0.864 	\\ \hline
% % 			R-BertScore
% % 			&  0.868 & \textbf{0.885} & 0.857 	\\ \hline
% 			F-BertScore
% 			&  0.868 & \textbf{0.879} & 0.861 	\\ \hline
% 		\end{tabular}
% 	\end{center}
%  	%\vspace{-10pt}
% 	\caption{Comparison of proposed Generation-Refinement Architecture for \pname\ with two Variants: 0 and 2 Refinement modules.}
% 	\label{tab:vcr_gen_refinement}
% 	%\vspace{-20pt}
% \end{wraptable}

\begin{figure}[t!]
\centering
\includegraphics[width=\linewidth]{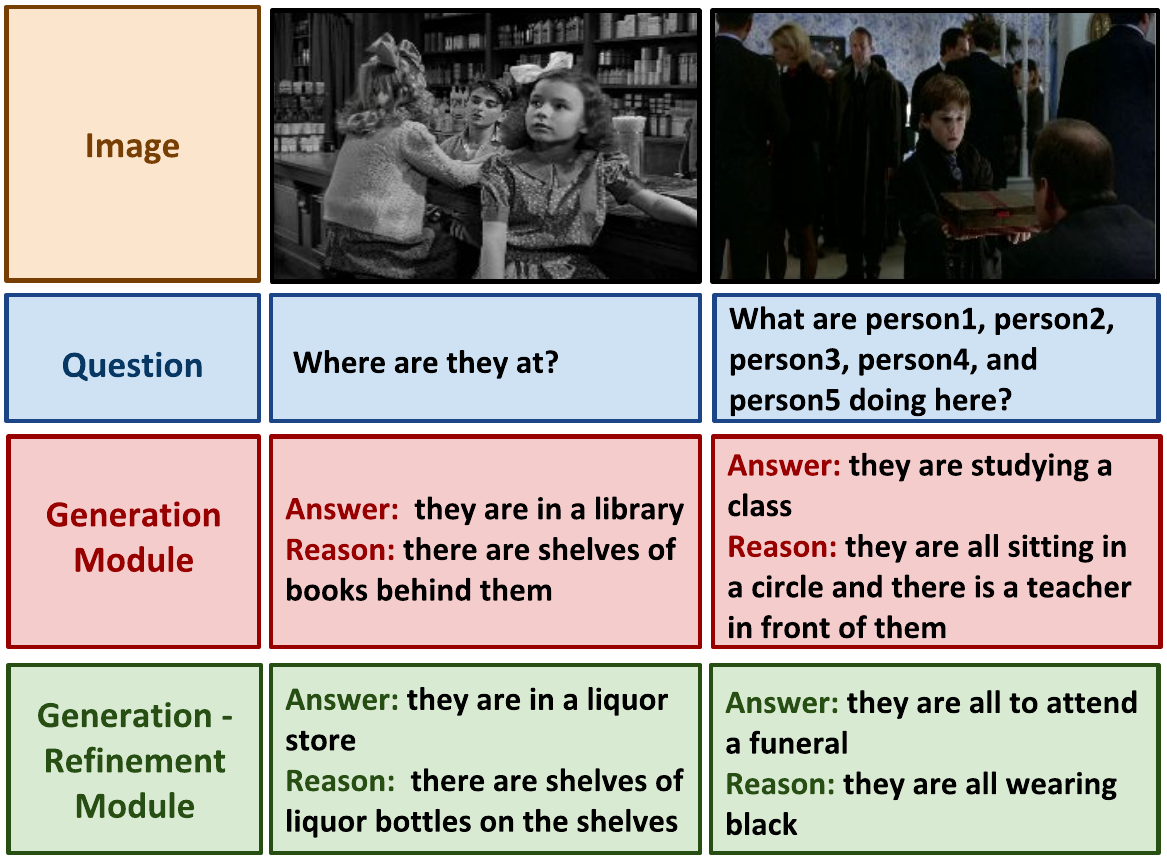}\label{fig:q1}
\caption{Qualitative results for our model with (in green, last row) and without (in red, penultimate row) Refinement module}
\label{fig:visual_vcr_qualitative_results}
\end{figure}

We hypothesize that the additional parameters (in a second Refinement module) in the model makes it harder for the network to learn. Removal of the refinement module also causes performance to drop, supporting our claim on the usefulness for a Refinement module too. Figure ~\ref{fig:visual_vcr_qualitative_results} provides a few qualitative results with and without the refinement module, supporting our claim. 
More results are presented in the supplementary.
% \sout{More qualitative results are presented in the supplementary.}
%Further, table~\ref{tab:vcr_classi_refinement} shows the classification accuracy of our proposed framework and model variants on the VCR dataset.
% \begin{figure*}[hbt]
% \centering
% %\vspace{8pt}
% \includegraphics[width=0.88\linewidth]{journal/figures/refinement_2_.pdf}\label{fig:q1}
% %\vspace{-6pt}
% \caption{Qualitative results for our model with (in green, last row) and without (in red, penultimate row) Refinement module}
% \label{fig:visual_vcr_qualitative_results}
% \end{figure*}

\paragraph{Transfer to other datasets.}
We also studied whether the proposed model, trained on the VCR dataset, can provide answers and rationales to visual questions in other VQA datasets (which do not have ground truth rationale provided). To this end, we tested our trained model on the widely used Visual7W \cite{Zhu2015Visual7WGQ} dataset without any additional training. 

\begin{figure}[!t]
\centering
\includegraphics[width=\linewidth]{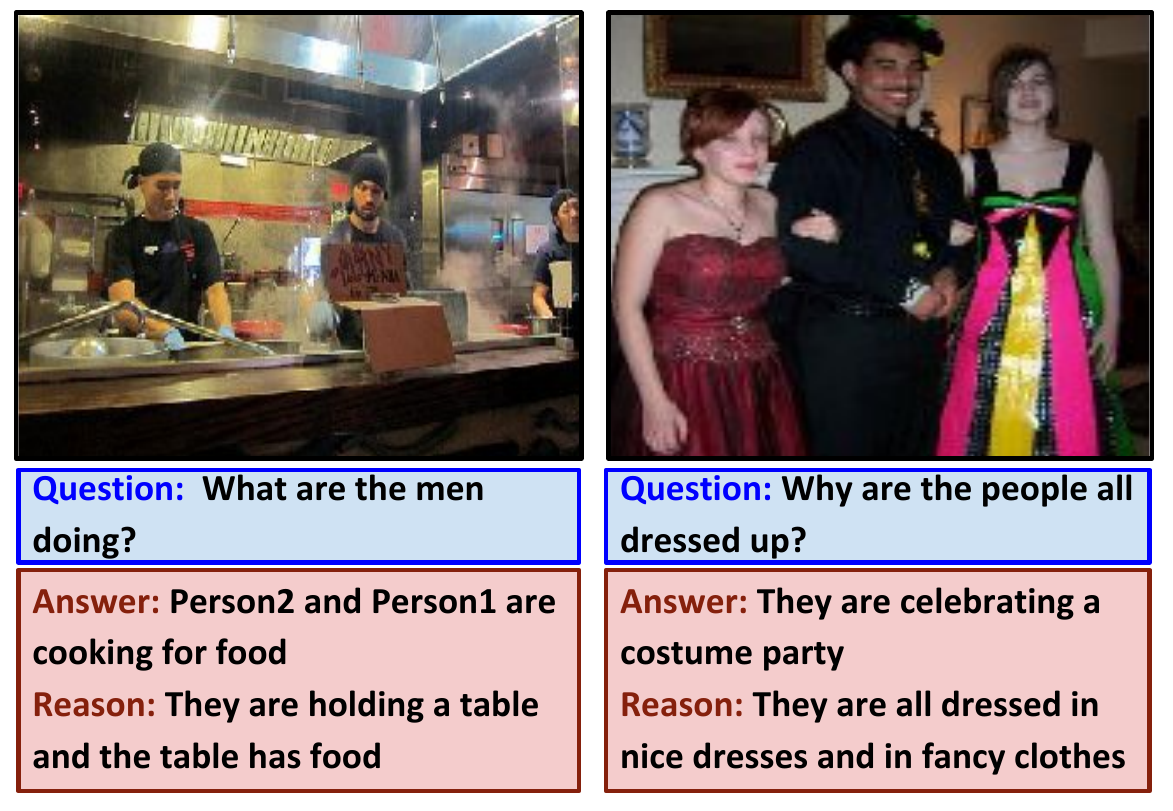}\label{fig:q1}
% \subfloat{\includegraphics[width=0.44\linewidth,height=5.3cm]{journal/figures/7w_Results_main2c_1_.pdf}\label{fig:q2}}
\caption{Qualitative results on Visual7W dataset (note that there is no rationale provided in this dataset, and all above rationales were generated by our model)}
\label{fig:visual_7w_qualitative_results}
\end{figure}

Figure \ref{fig:visual_7w_qualitative_results} presents qualitative results for \pname\ task on the Visual7W dataset. We also perform a Turing test on the generated answers and rationales to evaluate the model's performance on Visual7W. Thirty human evaluators were presented each with twenty five hand-picked image-question pairs, each of which contains a generated answer to the question and its rationale. The results, presented in Table \ref{tab:turing_test} show that our algorithm generalizes reasonably well to another VQA dataset and generates answers and rationales relevant to the image-question pair, without any explicit training for this dataset. This adds a promising dimension to this work. More results are presented in the supplementary.

\pname\ is a completely generative task and objective evaluation is a challenge, as in any other generative method. For comprehensive evaluation, we use a suite of objective metrics typically used in related vision-language tasks, and perform a Human Turing Test on the generated sentences. We perform a detailed analysis in the supplementary and show that even our successful results (qualitatively speaking) may have low scores on objective evaluation metrics at times, since generated sentences may not match a ground truth sentence word-by-word. We hope that opening up this dimension of generated explanations will only motivate a better metric in the near future.
%\vspace{-16pt}
\section{Conclusion}
\label{sec:conc}
%\vspace{-9pt}
In this paper, we propose \pname, a novel task for generating a multi-word answer and a rationale given an image and a question. Our work aims to go beyond classical VQA by moving to a completely generative paradigm. To solve \pname, we present an end-to-end generation-refinement architecture which is based on the observation that answers and rationales are dependent on one another. We showed the promise of our model on the VCR dataset both qualitatively and quantitatively, and our human Turing test showed results comparable to the ground truth.  We also showed that this model can be transferred to tasks without ground truth rationale. We hope that our work will open up a broader discussion around generative answers in VQA and other deep neural network models in general.
%\vspace{-14pt}
%\section{Acknowledgements}
%\label{sec:conc}
% \vspace{4pt}

\noindent \textbf{Acknowledgements.} We acknowledge MHRD and DST, Govt of India, as well as Honeywell India for partial support of this project through the UAY program, IITH/005. We also thank Japan International Cooperation Agency and IIT Hyderabad for the generous compute support.%We thank the anonymous reviewers for their valuable feedback, as well as all our lab members for all the insightful discussions at several stages of the project that improved the presentation of this work.
{\small
\bibliographystyle{ieee_fullname}
\bibliography{main_cvpr}
}
\clearpage
\beginsupplement
\section*{Supplementary Material: Beyond VQA: Generating Multi-word Answer and Rationale to Visual Questions}
In this supplementary section, we provide additional supporting information including:
% \vspace{-6pt}
\begin{itemize}
%\setlength\itemsep{-0.02em}
    % \item A high-level conceptual illustration of our proposed model (extension of Section 4 of main paper)
    % \item Dataset justification
    % \item Implementation details
    \item Results on VCR classification task (extension of Section 5 of main paper)
    \item Additional qualitative results of our model (extension of Section 5 of main paper)
    % \item Additional details on conducting Human Turing test (extension of section 5 of our paper)
    \item Qualitative results on impact of the Refinement module in our model, and a study on the effect of adding refinement module to VQA-E
    % \item Comparison between the model with and without refinement module
    \item Qualitative results on transferring our model to the VQA task (extension of results in Section 5 of main paper)
    % \item More Qualitative Results on Visual7W dataset
    \item A discussion on existing objective evaluation metrics for this task, and the need to go beyond (extension of section 5 of our paper)
    % \item A study, using our attention maps, on the interpretability of our results
    % \item Interpretability of proposed end-to-end generation refinement architecture by exploiting visual attention maps 
\end{itemize}
\section{VCR classification task}
\begin{table*}[t!]
\footnotesize
\vspace{-2pt}
\centering
\begin{adjustbox}{max width=\textwidth}
\begin{tabular}{|l|c|c|c|c|c|c|c|c|c|}
\hline
\multicolumn{1}{|c|}{\textbf{Metrics}} & \multicolumn{3}{c|}{\textbf{Q+I+C}}                     & \multicolumn{3}{c|}{\textbf{Q+I}}                       & \multicolumn{3}{c|}{\textbf{Q+C}}                       \\ \hline
                                       & \textbf{Answer} & \textbf{Rationale} & \textbf{Overall} & \textbf{Answer} & \textbf{Rationale} & \textbf{Overall} & \textbf{Answer} & \textbf{Rationale} & \textbf{Overall} \\ \hline
\textbf{Infersent}                     & 34.90           & 31.78              & 11.91            & 34.73           & 31.47              & 11.68            & 30.50           & 27.99              & 9.17             \\ \hline
\textbf{USE}                           & 34.56           & 30.81              & 11.13            & 34.7            & 30.57              & 11.17            & 30.15           & 27.57              & 8.56             \\ \hline
\end{tabular}
\end{adjustbox}
\caption{Quantitative results on the VCR dataset. Accuracy percentage for answer classification, rationale classification and overall answer-rationale classification is reported.}
\label{tab:vcr_classification_accuracy}
\vspace{-8pt}
\end{table*}
We evaluate the performance of our model on the classification task. For every question, there are four answer choices and four rationale choices provided in the VCR dataset. We compute the similarity scores between each of the options and our generated answer/ rationale, and choose the option with the highest similarity score. 
% If it gets either the answer or the rationale wrong, the entire prediction will be wrong.
% \sout{Only samples which correctly predict \textit{both} answers and rationales are considered, and accuracy percentage is reported in Table \ref{tab:vcr_classi_best}.
% However, we report only answer and only rationale classification accuracy in the supplementary.}
Accuracy percentage for answer classification, rationale classification and overall answer+rationale classification (denoted as Overall) are reported in Table \ref{tab:vcr_classification_accuracy}. Only samples that correctly predict \textit{both} answers and rationales are considered for overall answer+rationale classification accuracy. The results show the difficulty of the \pname task, expounding the need for opening up this problem to the community.
\input\section{Additional qualitative results}
\begin{figure*}
\centering
\subfloat{\includegraphics[width=0.49\linewidth,height=2.6cm]{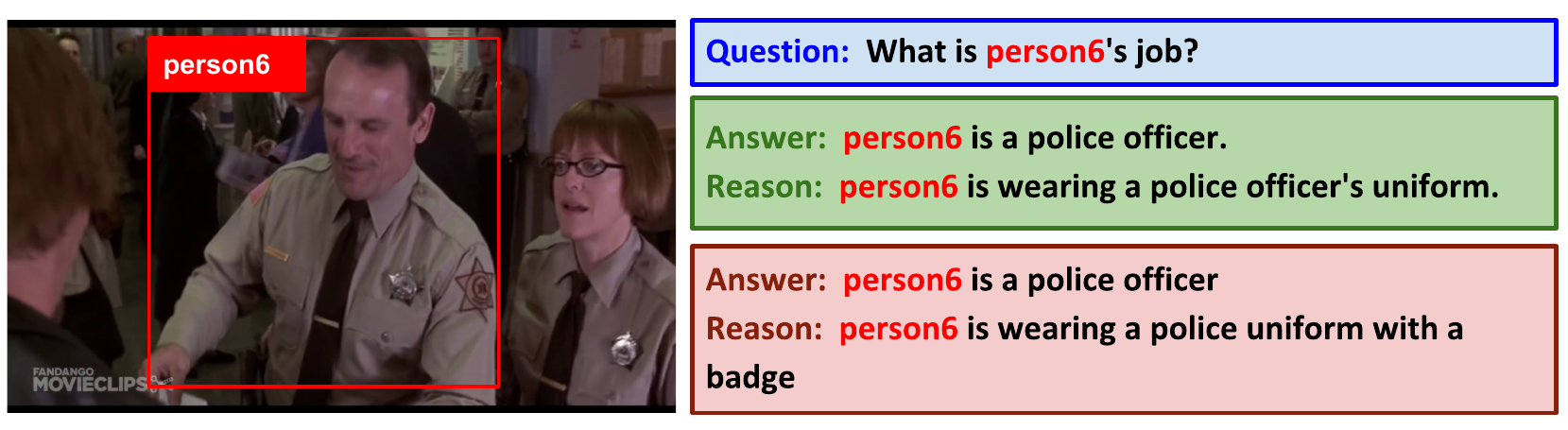}\label{fig:q1}}
\subfloat{\includegraphics[width=0.49\linewidth,height=2.6cm]{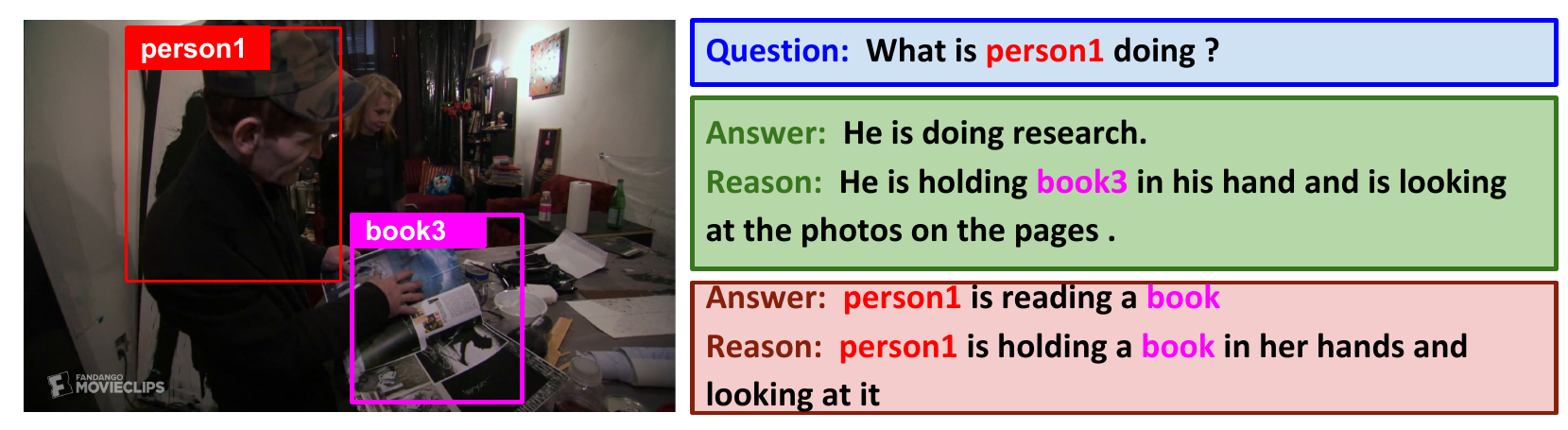}}\\ \vfill
\subfloat{\includegraphics[width=0.49\linewidth,height=2.6cm]{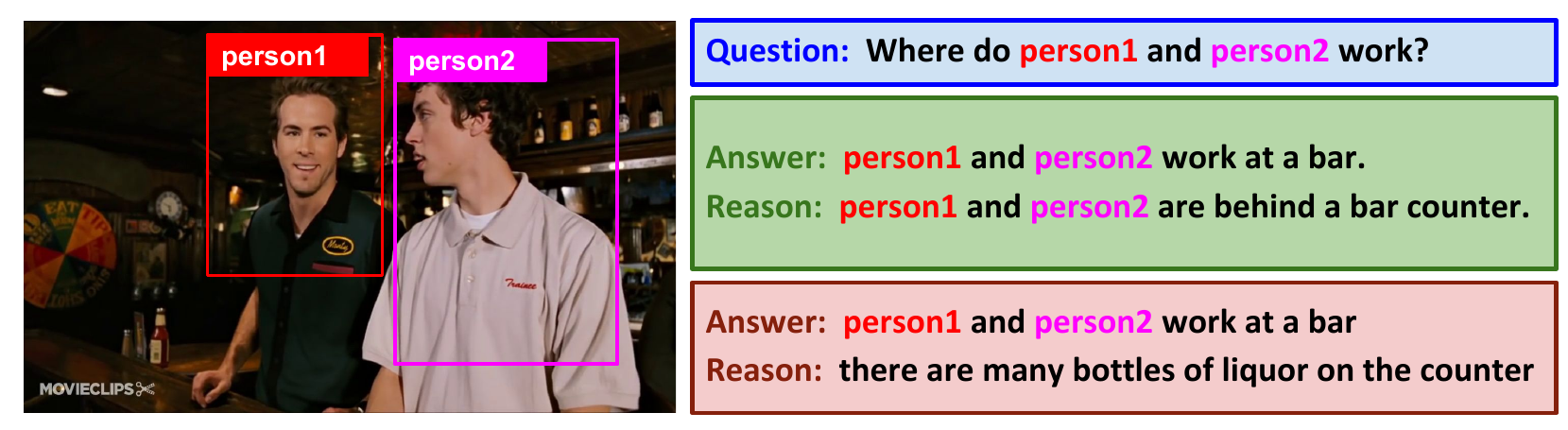}}
\subfloat{\includegraphics[width=0.49\linewidth,height=2.6cm]{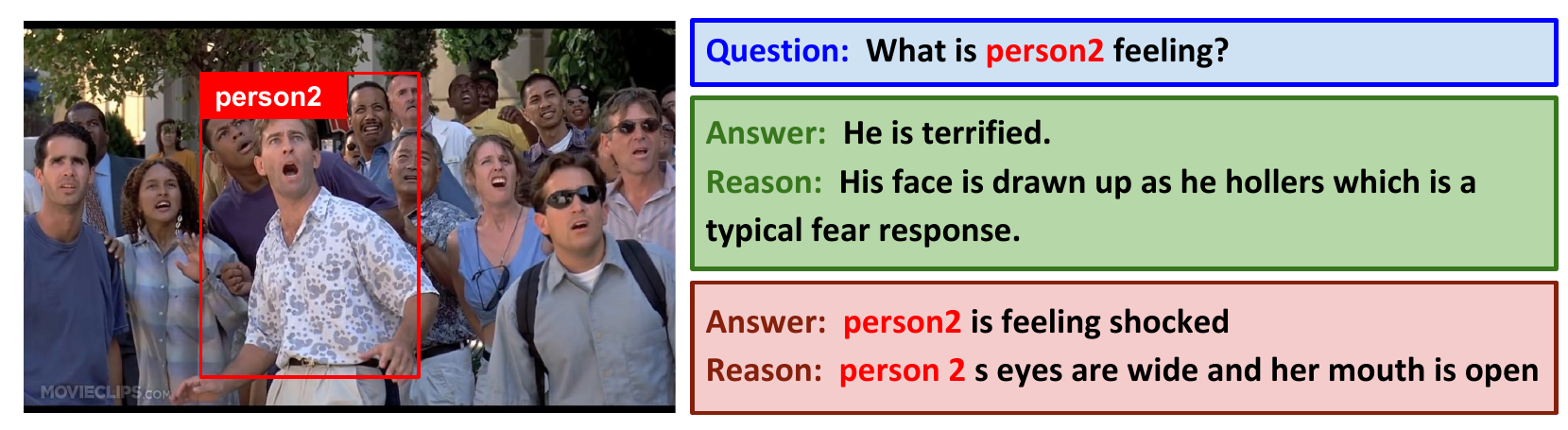}}\\ \vfill
\subfloat{\includegraphics[width=0.49\linewidth,height=2.6cm]{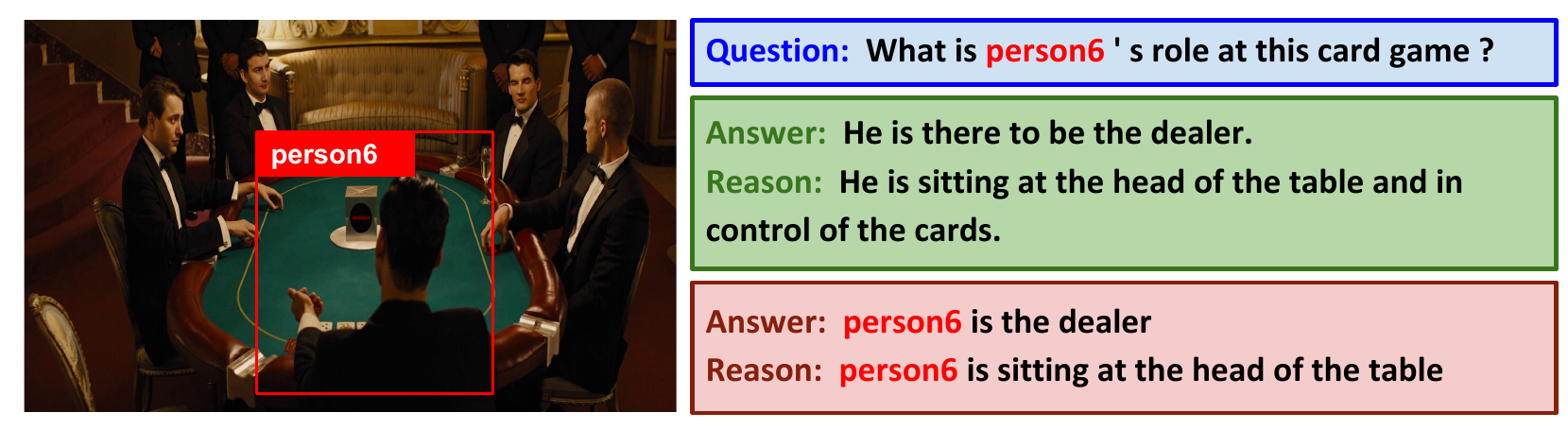}}
\subfloat{\includegraphics[width=0.49\linewidth,height=2.6cm]{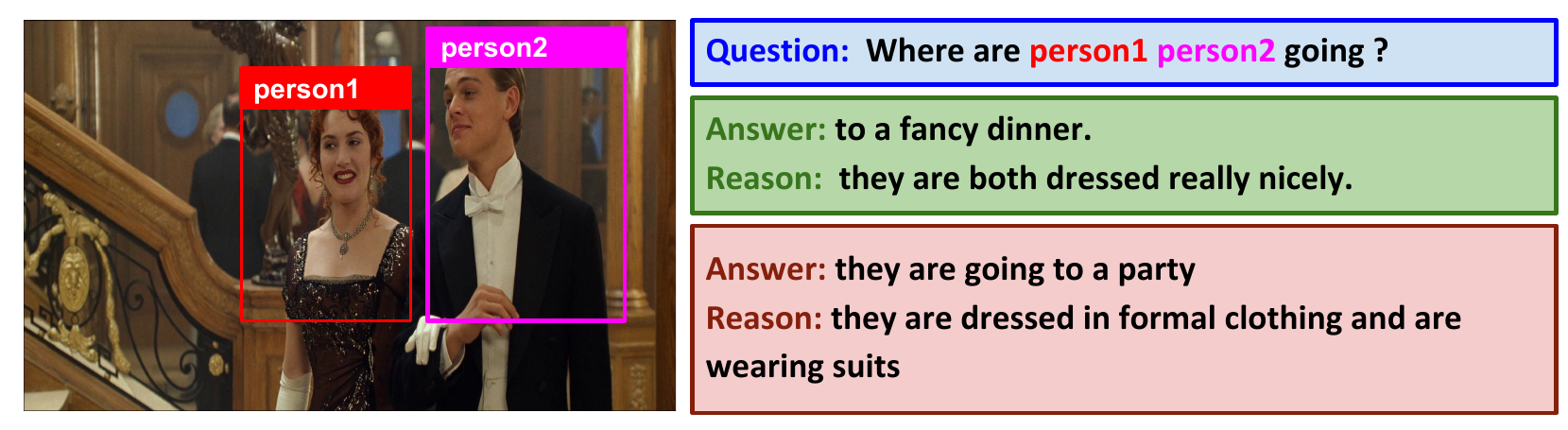}}\\ \vfill
% \subfloat{\includegraphics[width=0.49\linewidth]{appendix/figures_appendix/qual_results/Supp_Results7(26).pdf}\label{fig:q1}}
\subfloat{\includegraphics[width=0.49\linewidth,height=2.6cm]{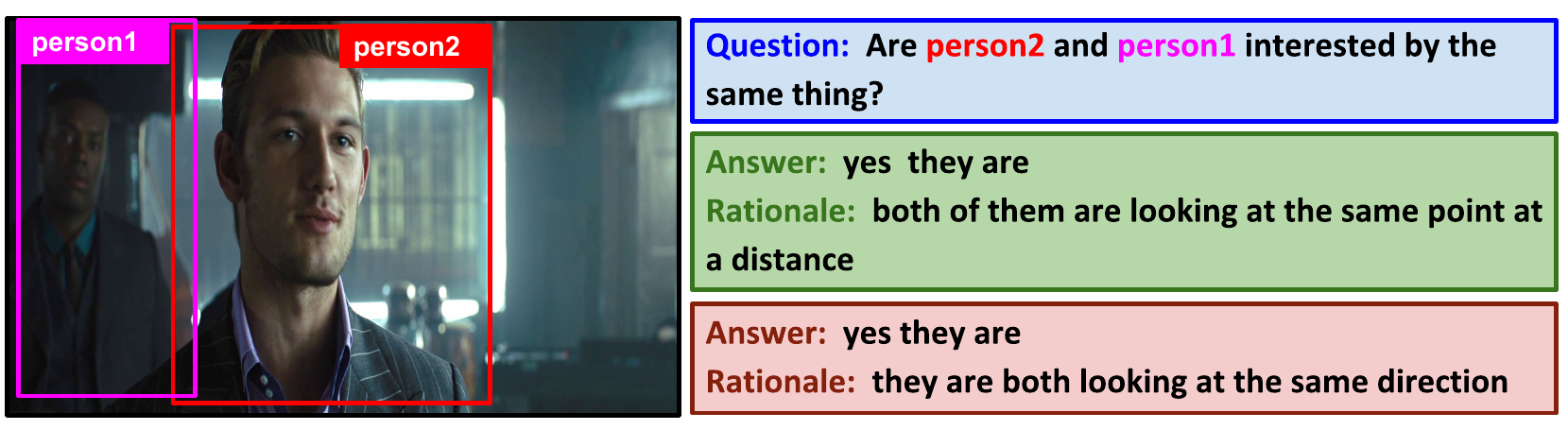}}
\subfloat{\includegraphics[width=0.49\linewidth,height=2.6cm]{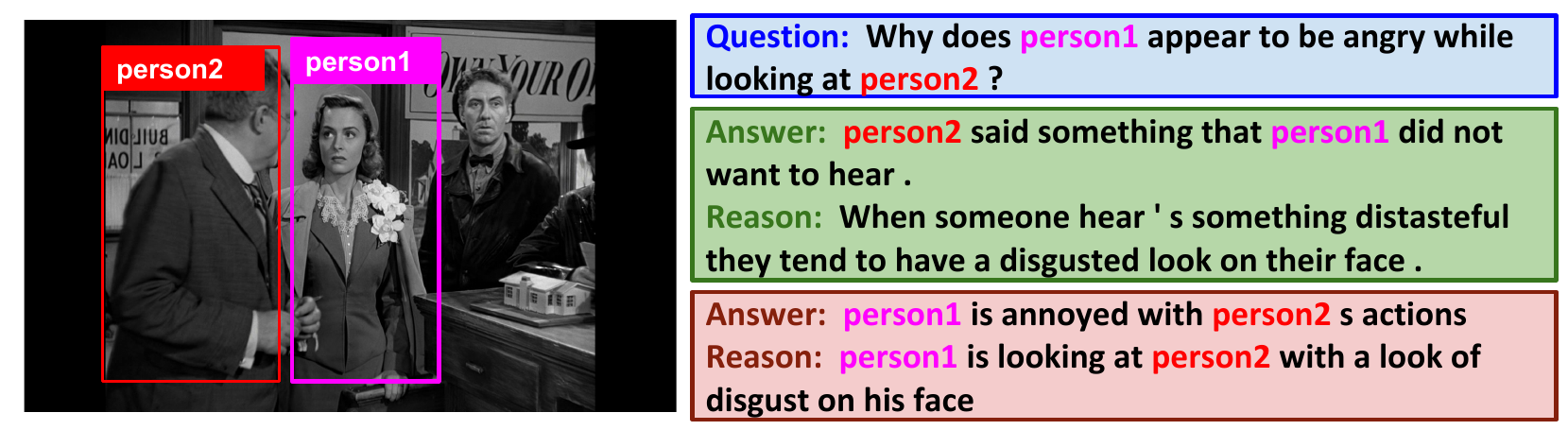}}\\ \vfill
\subfloat{\includegraphics[width=0.49\linewidth,height=2.6cm]{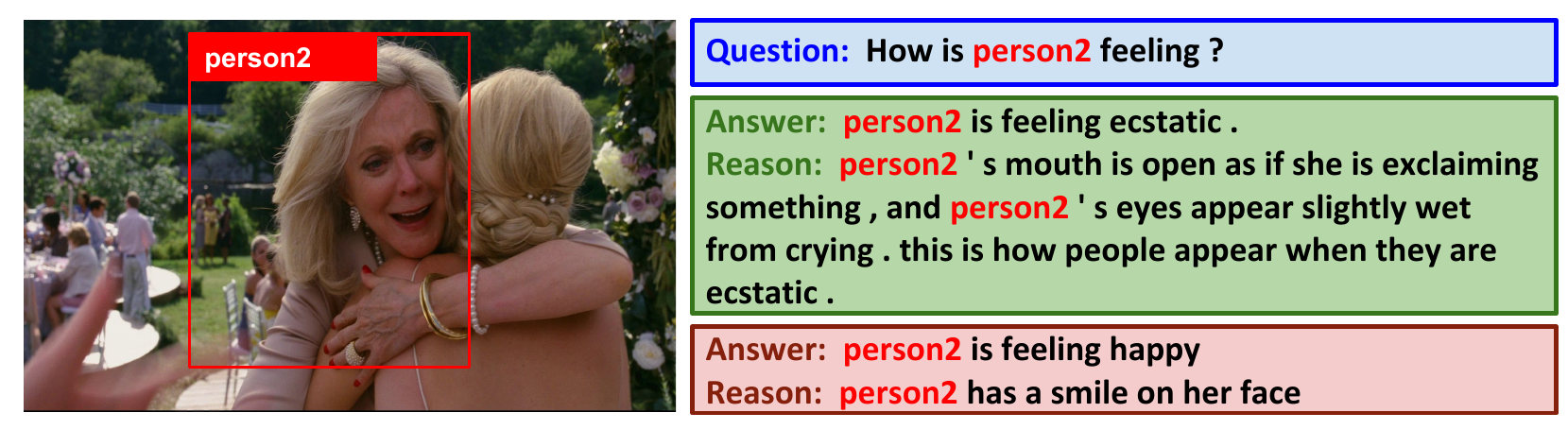}}
\subfloat{\includegraphics[width=0.49\linewidth,height=2.6cm]{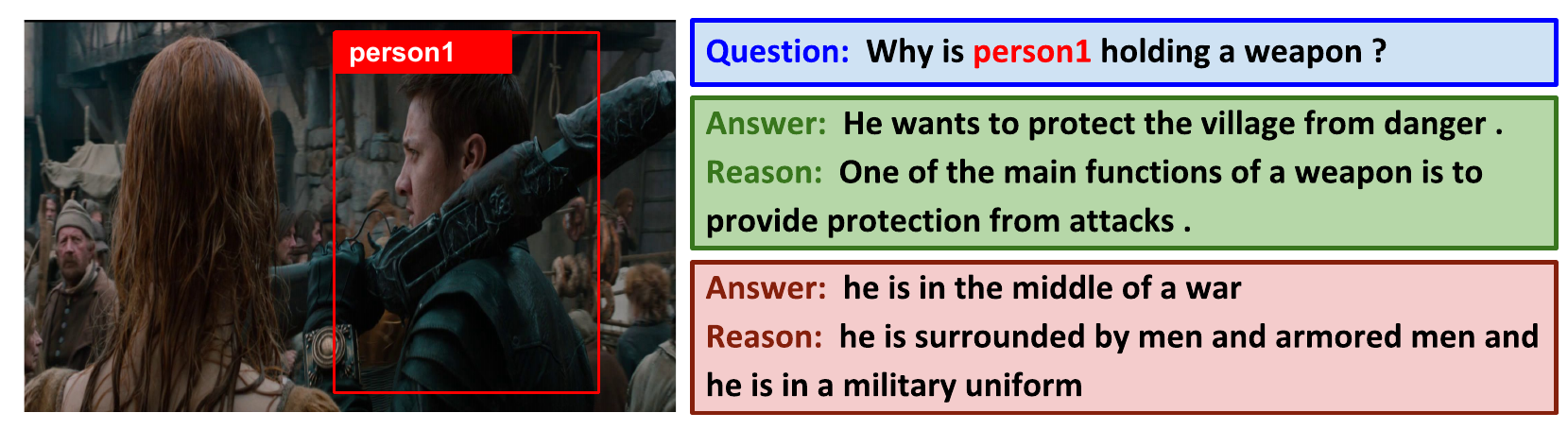}}\\ \vfill
\subfloat{\includegraphics[width=0.49\linewidth,height=2.6cm]{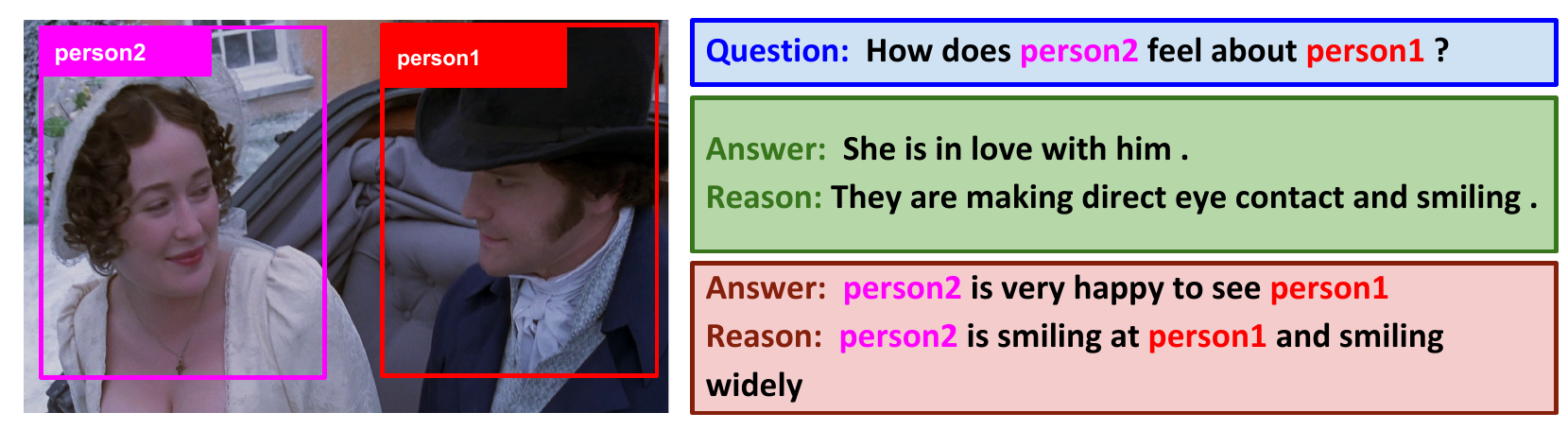}}
\subfloat{\includegraphics[width=0.49\linewidth,height=2.6cm]{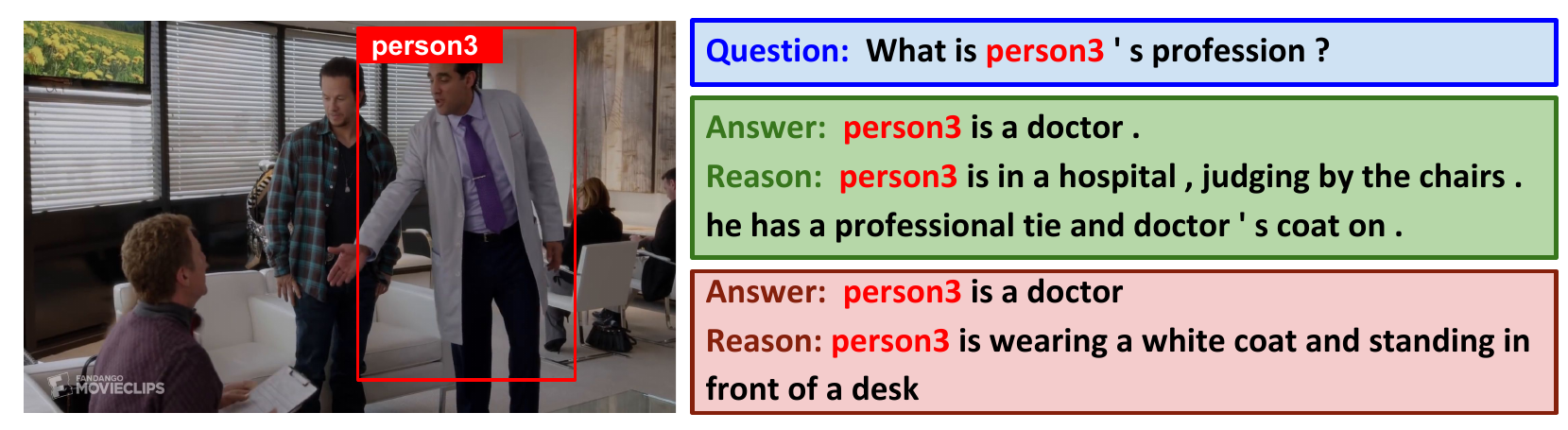}}\\ \vfill
% \subfloat{\includegraphics[width=0.49\linewidth,height=2.6cm]{appendix/figures_appendix/qual_results/Supp_Results5_25139_.pdf}\label{fig:q1}}
% \subfloat{\includegraphics[width=0.49\linewidth,height=2.6cm]{appendix/figures_appendix/qual_results/Supp_Results6_26293_.pdf}\label{fig:q2}}\\
% \subfloat{\includegraphics[width=0.49\linewidth,height=2.6cm]{figures/Results_2_2401_.pdf}\label{fig:q2}}
% \subfloat{\includegraphics[width=0.49\linewidth,height=2.6cm]{appendix/figures_appendix/qual_results/Supp_Results8_26002_.pdf}\label{fig:q2}}\\
\caption{\textit{(Best viewed in color)} Qualitative results for \pname\ task from our Generation Refinement architecture. Blue box = question about image; Green = Ground truth; Red = Generated results from our proposed architecture. (Object regions shown on image is for reader's understanding and are not given as input to model.)}
% \vspace{-12pt}
\label{fig:more_qualitative_results}
\end{figure*}

\begin{figure*}[hbt]
\centering
\subfloat{\includegraphics[width=0.49\linewidth,height=2.6cm]{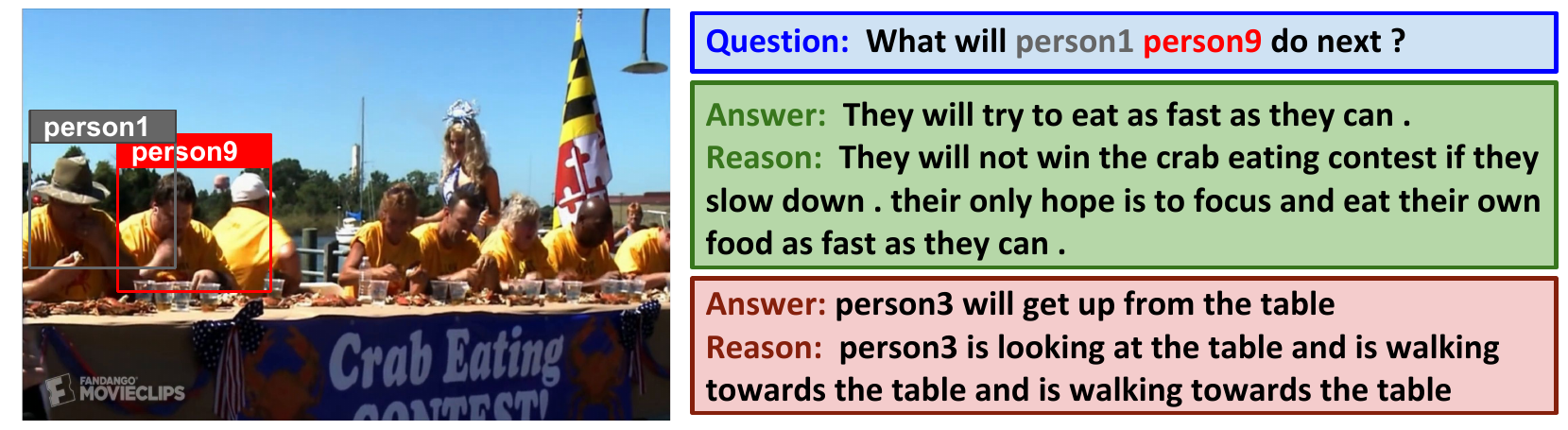}\label{fig:q2}}
\subfloat{\includegraphics[width=0.49\linewidth,height=2.6cm]{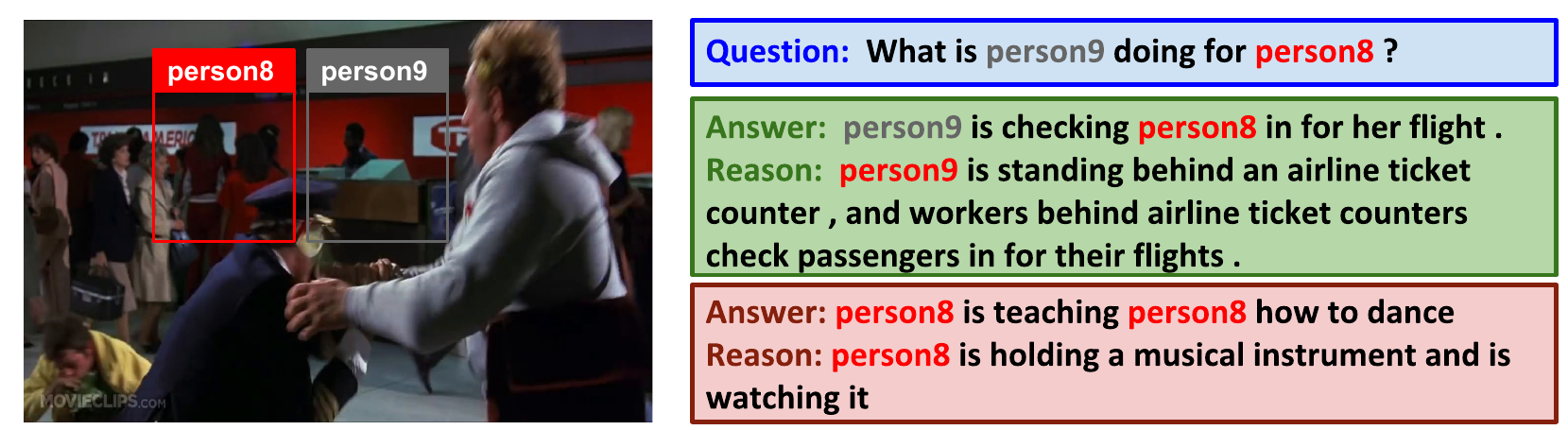}}\\
\subfloat{\includegraphics[width=0.49\linewidth,height=2.6cm]{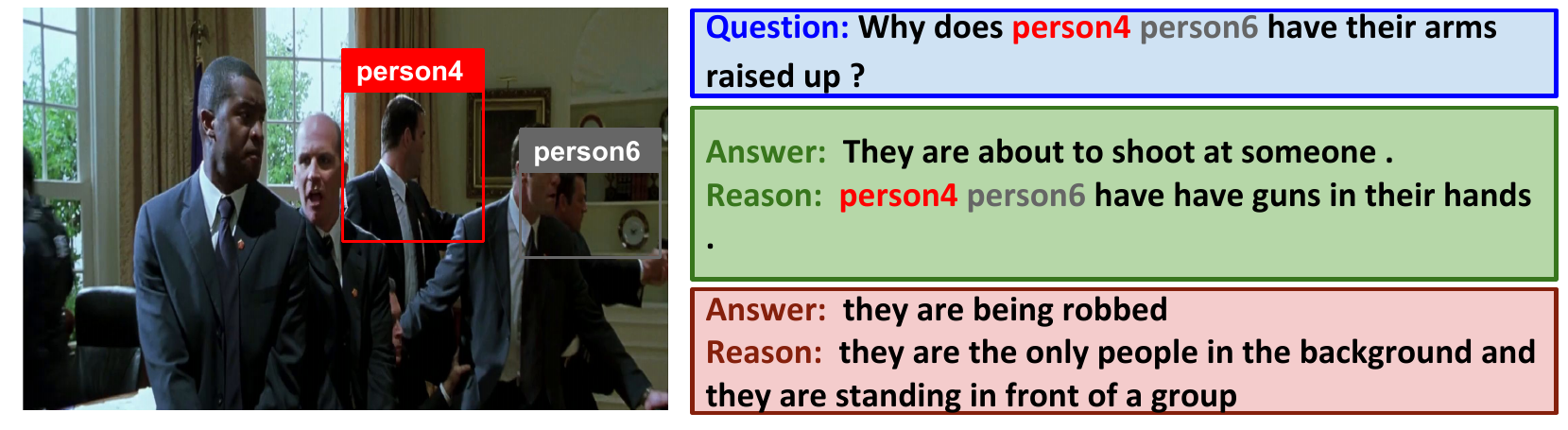}}
\subfloat{\includegraphics[width=0.49\linewidth,height=2.6cm]{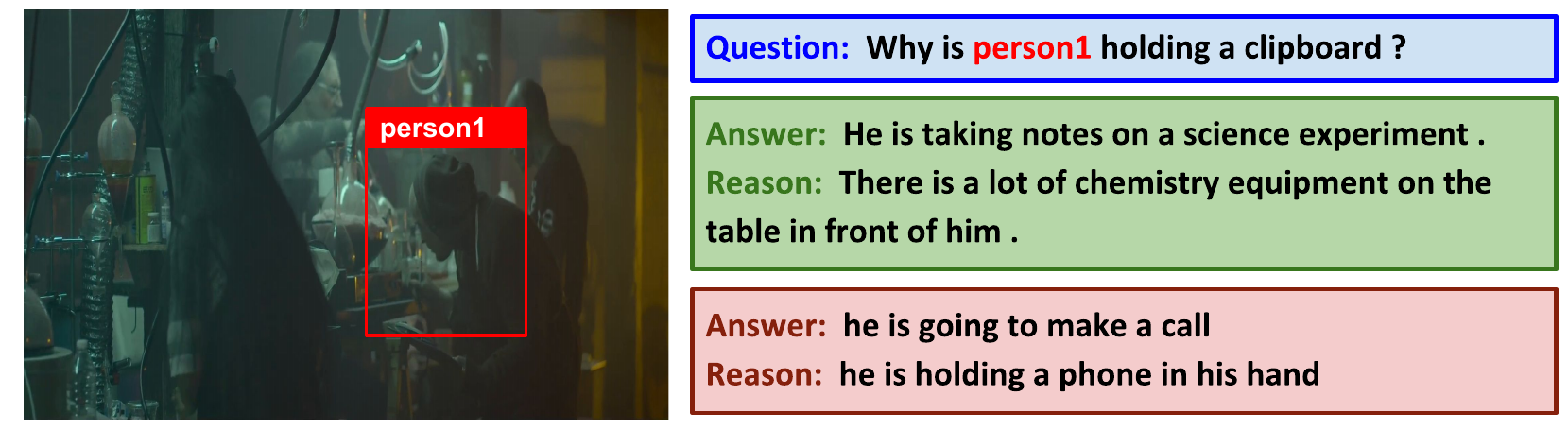}}\\
\subfloat{\includegraphics[width=0.49\linewidth,height=2.6cm]{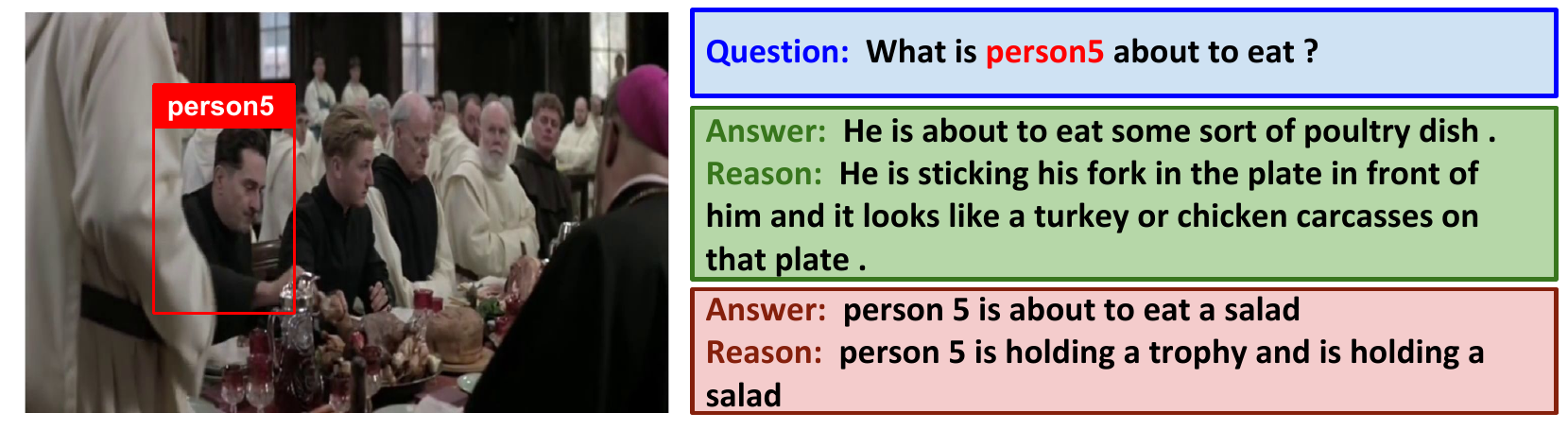}}
\subfloat{\includegraphics[width=0.49\linewidth,height=2.6cm]{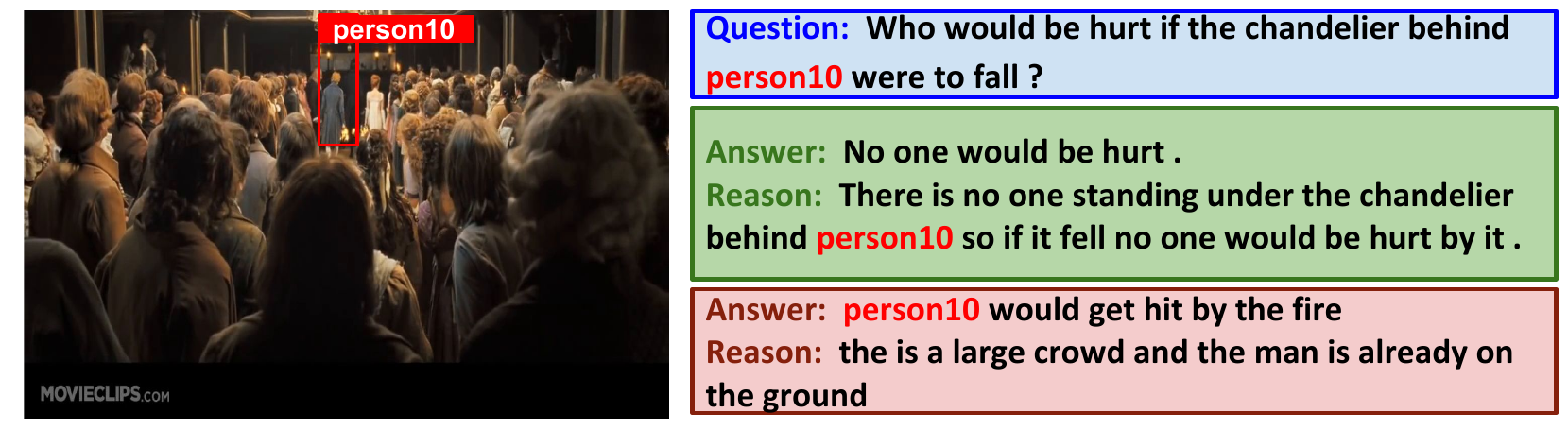}}\\
\subfloat{\includegraphics[width=0.49\linewidth,height=2.6cm]{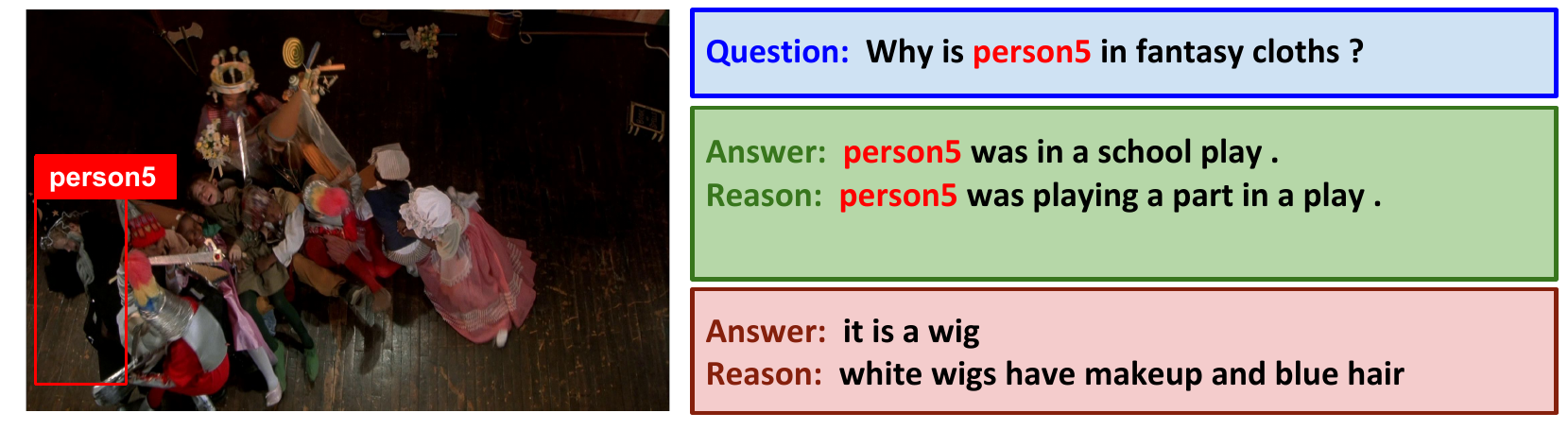}}
\subfloat{\includegraphics[width=0.49\linewidth,height=2.6cm]{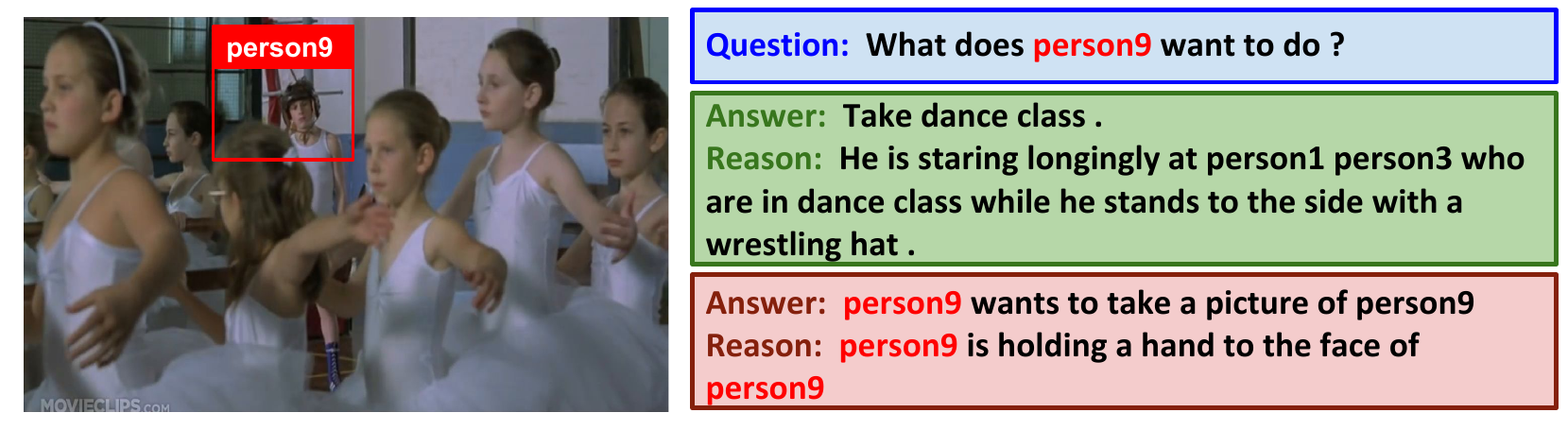}}\\
\caption{\textit{(Best viewed in color)} Challenging examples for which our model fails to generate the semantically correct answer and rationale. Blue box = question about image; Green = Ground truth; Red = Generated results from our proposed architecture. (Object regions shown on image are for reader's understanding and are not given as input to model.)}
% \vspace{-12pt}
\label{fig:neg_qualitative_results}
\end{figure*}

In addition to the qualitative results presented in Section 5 of the main paper, Figure~\ref{fig:more_qualitative_results} presents more qualitative results from our proposed model on the VCR dataset for the \pname\ task. We observe that our model is capable of generating answer-rationale pairs to complex subjective questions of the type: Explanation (why, how come), Activity (doing, looking, event), Temporal (happened, before, after, etc), Mental (feeling, thinking, love, upset), Scene (where, time) and Hypothetical sentences (if, would, could).
For completeness of understanding, we also show a few more examples on which our model fails to generate a good answer-rationale pair in Figure~\ref{fig:neg_qualitative_results}. As stated earlier in Section 5, even on these results, we observe that our model does generate both answers and rationales that are grammatically correct and complete. Improving the semantic correctness of the generations will be an important direction of future work. 
% Figure~\ref{fig:neg_qualitative_results} presents a few examples from the VCR dataset on which our model fails to generate a good answer-rationale pair. We observe that the failure in these cases is mainly due to the obscure image (images with large number of objects)
% The question refers to that part of the image which is not even visible clearly.
% high complexity of the image, which makes it difficult for a model to answer subjective questions about the image.
\section{Impact of refinement module}

% \begin{figure*}
% \centering
%     \subfloat{\includegraphics[width=\linewidth]{appendix/figures_appendix/attention/atten-part2.pdf}\label{fig:q1}}\\[-2ex]
%     \subfloat{\includegraphics[width=\linewidth]{appendix/figures_appendix/attention/atten2-part2_1.pdf}\label{fig:q2}}\\[-2ex]
%     \subfloat{\includegraphics[width=\linewidth]{appendix/figures_appendix/attention/atten1-part2.pdf}\label{fig:q3}}
% \caption{\textit{(Best viewed in color)} Visualization of attention weights that models use while generating each word.}
% \label{fig:attention maps2}
% \end{figure*}

\begin{figure*}
\centering
    \subfloat{\includegraphics[width=0.92\linewidth]{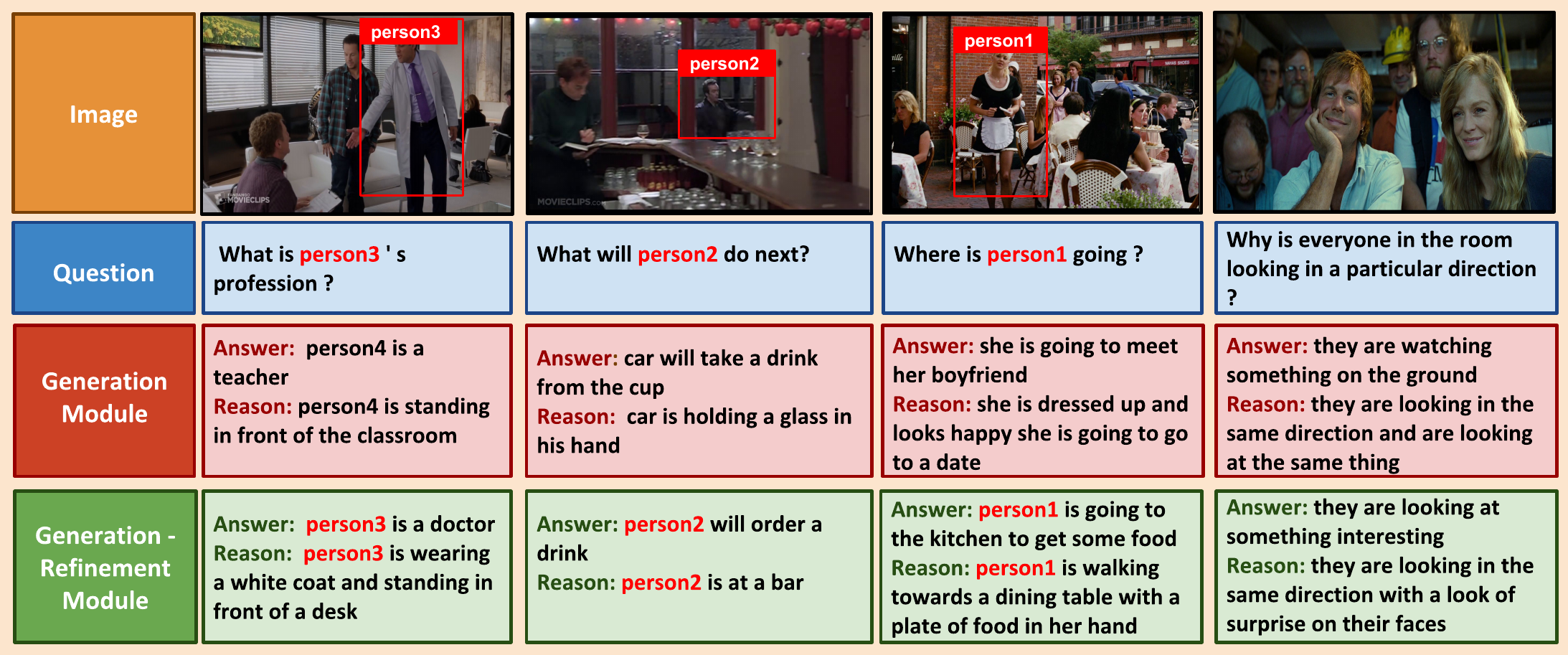}\label{fig:q3}}\\
    \subfloat{\includegraphics[width=0.92\linewidth]{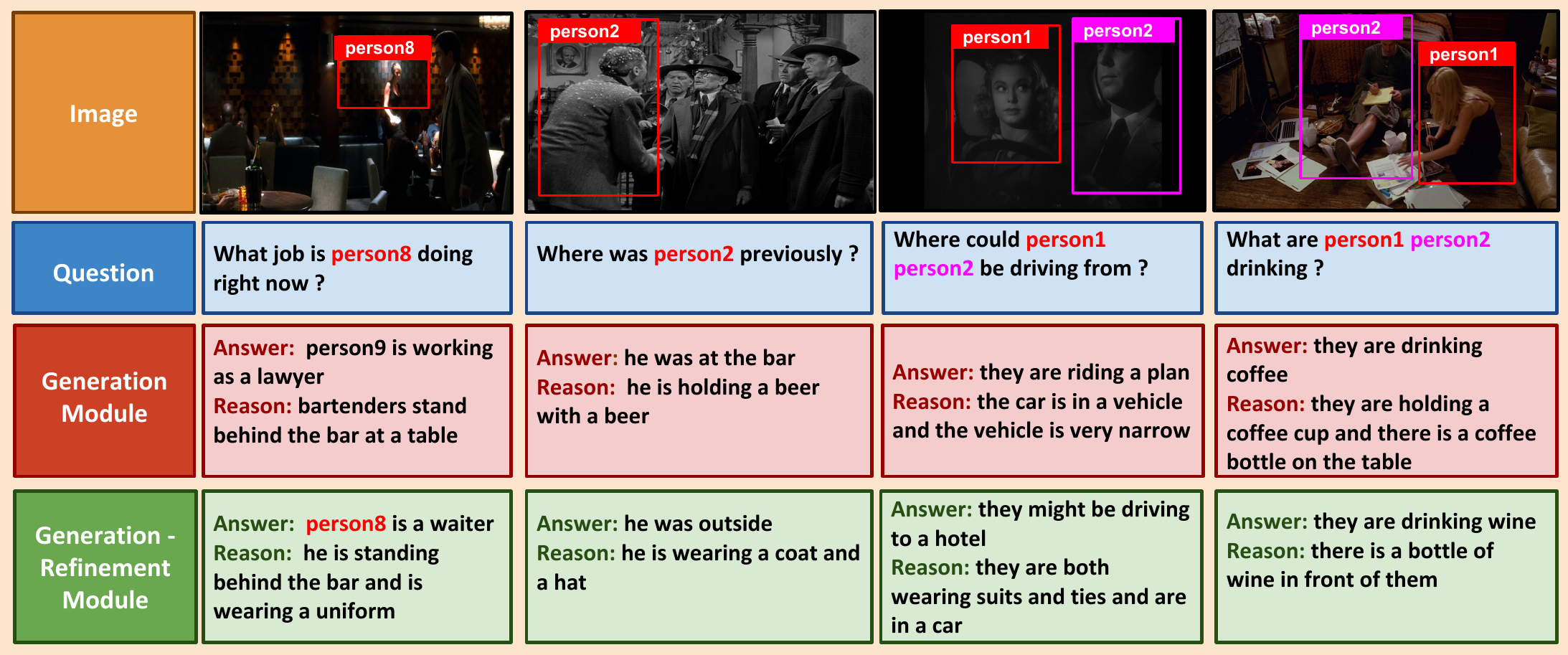}}\\
    \subfloat{\includegraphics[width=0.92\linewidth]{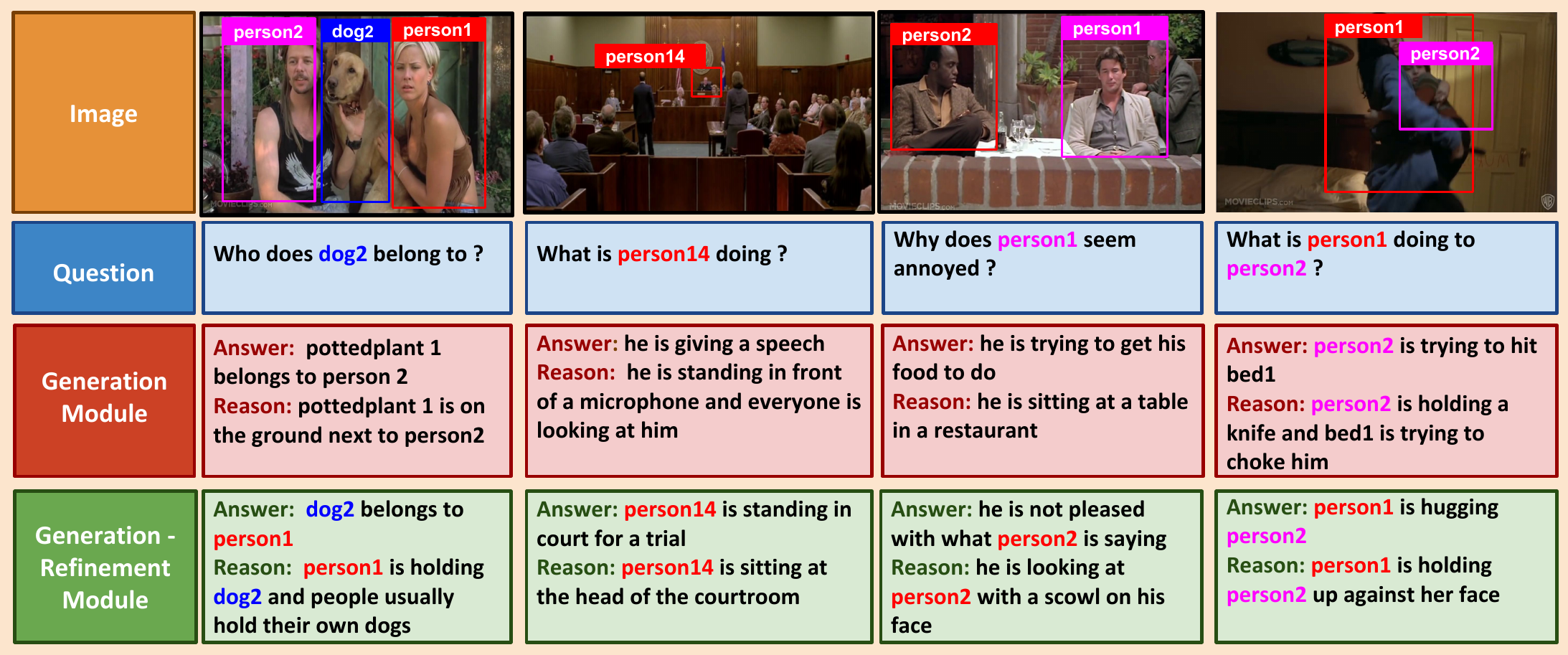}}
\caption{\textit{(Best viewed in color)} Qualitative results for the model with and without Refinement module. Blue box = question about image; Green  = Results from model with Refinement module; Red = Results from model without Refinement module. (Object regions shown on image are for reader's understanding and are not given as input to model.)}
\label{fig:qualitative_results_ref}
\vspace{-2pt}
\end{figure*}

 Figure~\ref{fig:qualitative_results_ref} provides a few examples to qualitatively compare the model with and without the refinement module, in continuation to the discussion in Section 6. We observe that the model without the refinement module fails to generate answers and rationale for complex image-question pairs. However, our proposed Generation-Refinement model is capable of generating a meaningful answer with a supporting explanation. Hence the addition of the refinement module to our model is useful to generate answer-rationale pairs to complex questions.
 %\section{A study on addition of refinement module to VQA Models}
We also performed a study on the effect of adding refinement to VQA models, particularly to VQA-E~\cite{Li2018VQAEEE}, which can be considered close to our work since it provides explanations as a classification problem (it classifies the explanation among a list of options, unlike our model which generates the explanation). To add refinement, we pass the last hidden state of the LSTM that generates the explanation along with joint representation to another classification module. %We intuit that explicit knowledge of the explanation may help answer classification. 
However, we did not observe improvement in classification accuracy when the refinement module is added for such a model. This may be attributed to the fact that the VQA-E dataset consists largely of one-word answers to visual questions. We infer that the interplay of answer and rationale, which is important to generate a better answer and provide justification, is more useful in multi-word answer settings which is the focus of this work. %This may be attributed to the fact that many questions in VQA are completely 'visual', in the sense that commonsense reasoning is not required to answer the question. Providing a post-hoc rationale to the classifier serves no purpose in this case, since the question can be answered correctly by just 'looking' at the image.  
%As shown in our work, refinement is most useful in the case of questions that require multi-word answers and commonsense reasoning, where there is an interplay between answer and rationale.
%  This supports the observation that answer and rationale are dependent on one another.
%  Our Generation-Refinement architecture which is based on the observation that answers and rationales are dependent on one another is capable of generating a meaningful answer and a reason to justify it.
%  Thus the generated reason from the rationale generator helps the answer refiner to generate a better answer.
\vspace{-8pt}
\section{Qualitative results on transfer to VQA task}
\begin{figure*}
\centering
\subfloat{\includegraphics[width=0.45\linewidth]{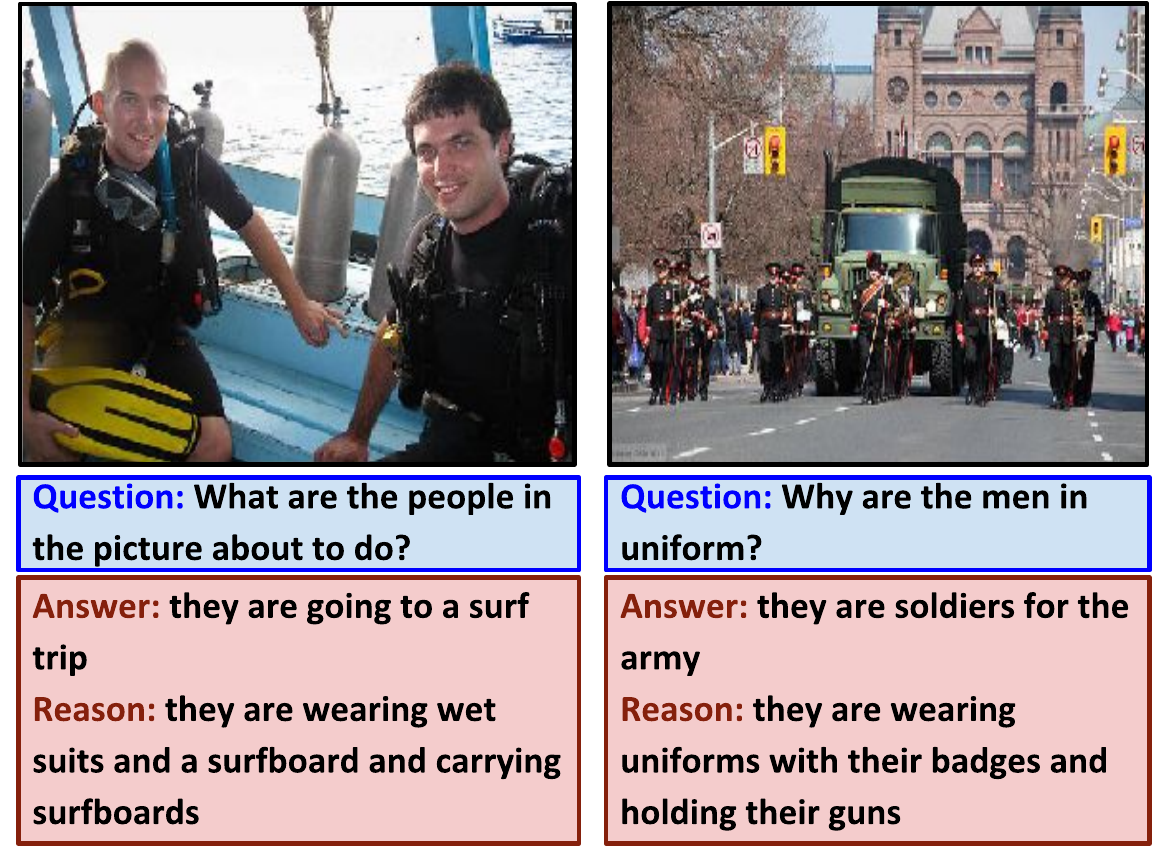}\label{fig:q1}}
\subfloat{\includegraphics[width=0.45\linewidth]{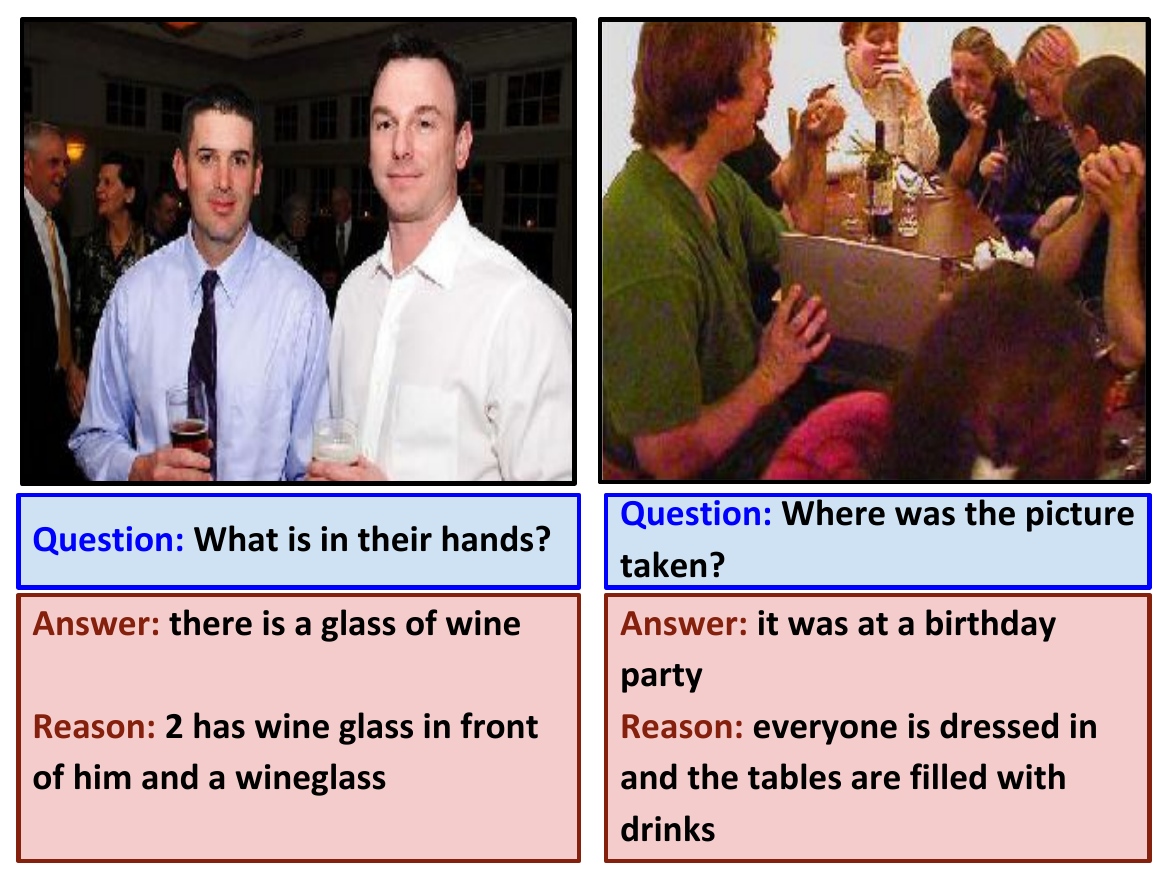}\label{fig:q2}}\hfill
\subfloat{\includegraphics[width=0.45\linewidth]{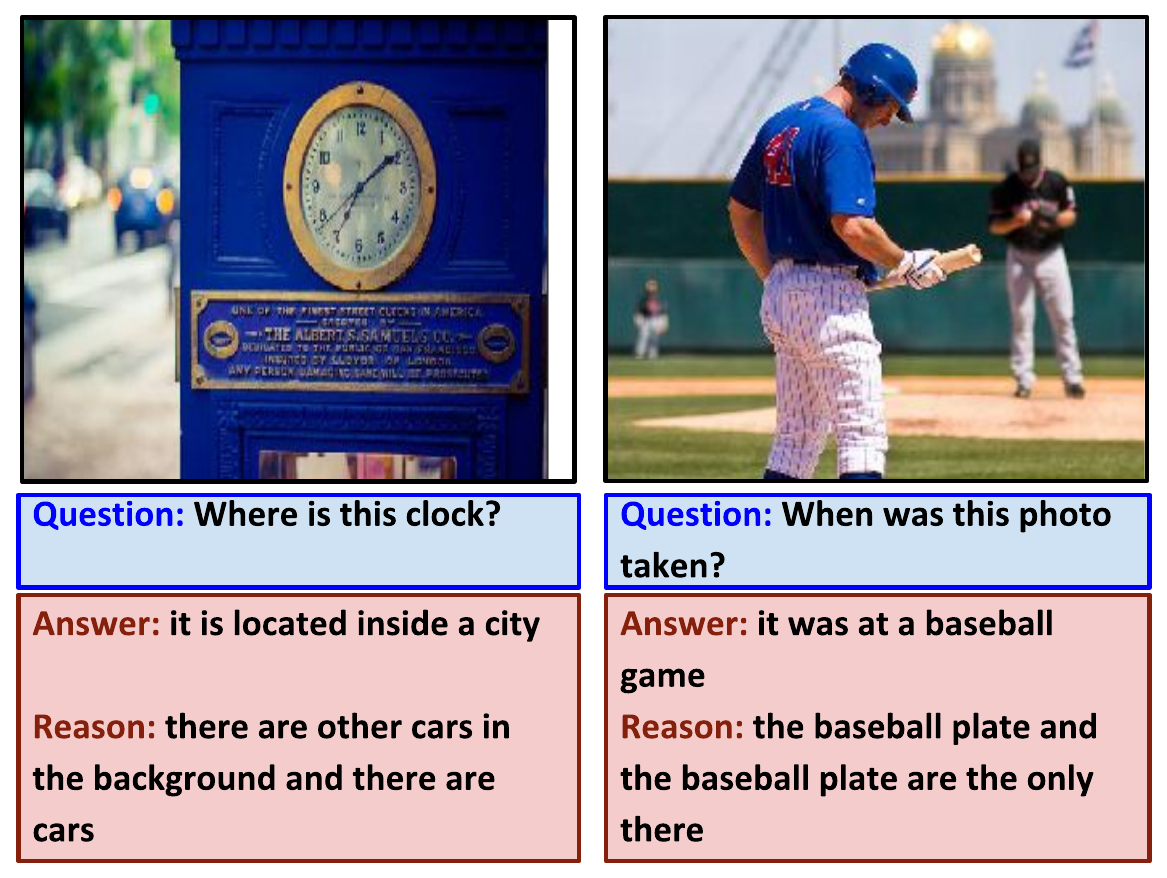}\label{fig:q1}}
\subfloat{\includegraphics[width=0.45\linewidth]{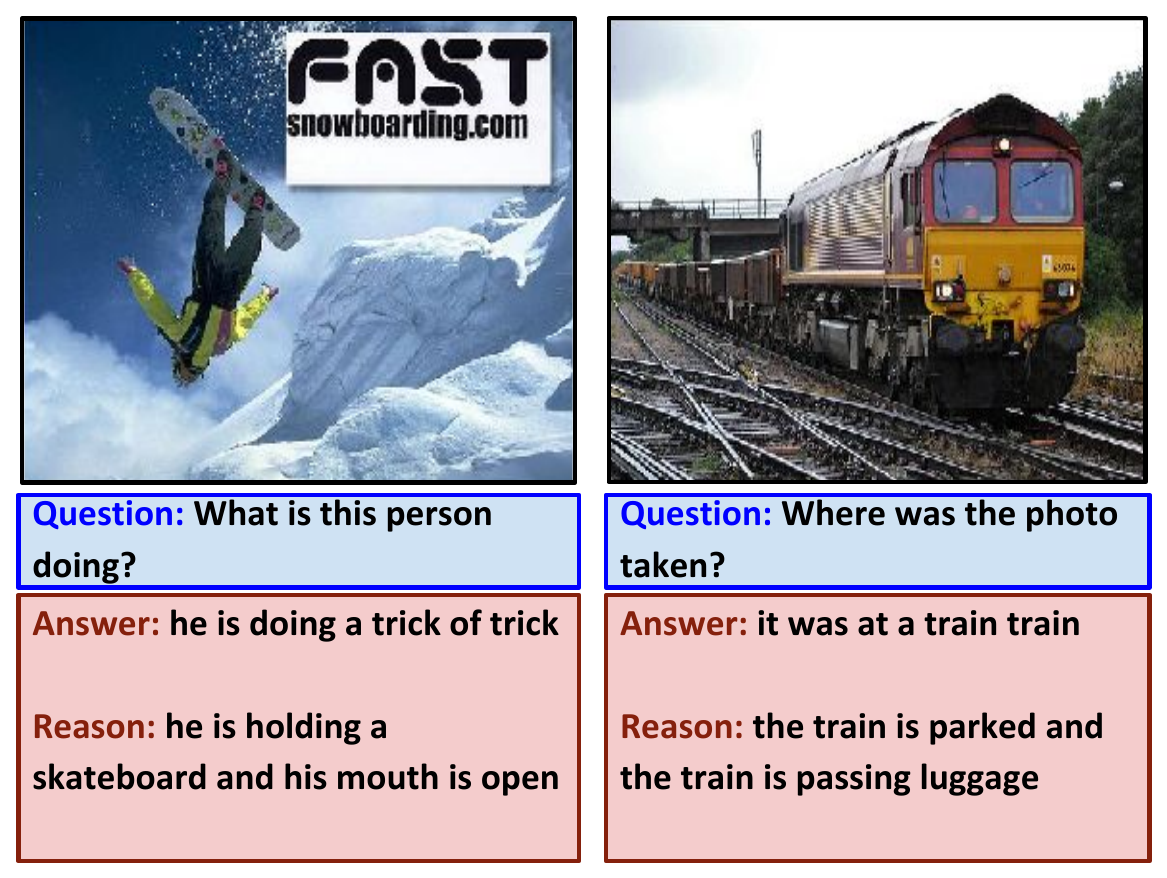}\label{fig:q2}}\hfill
\subfloat{\includegraphics[width=0.45\linewidth]{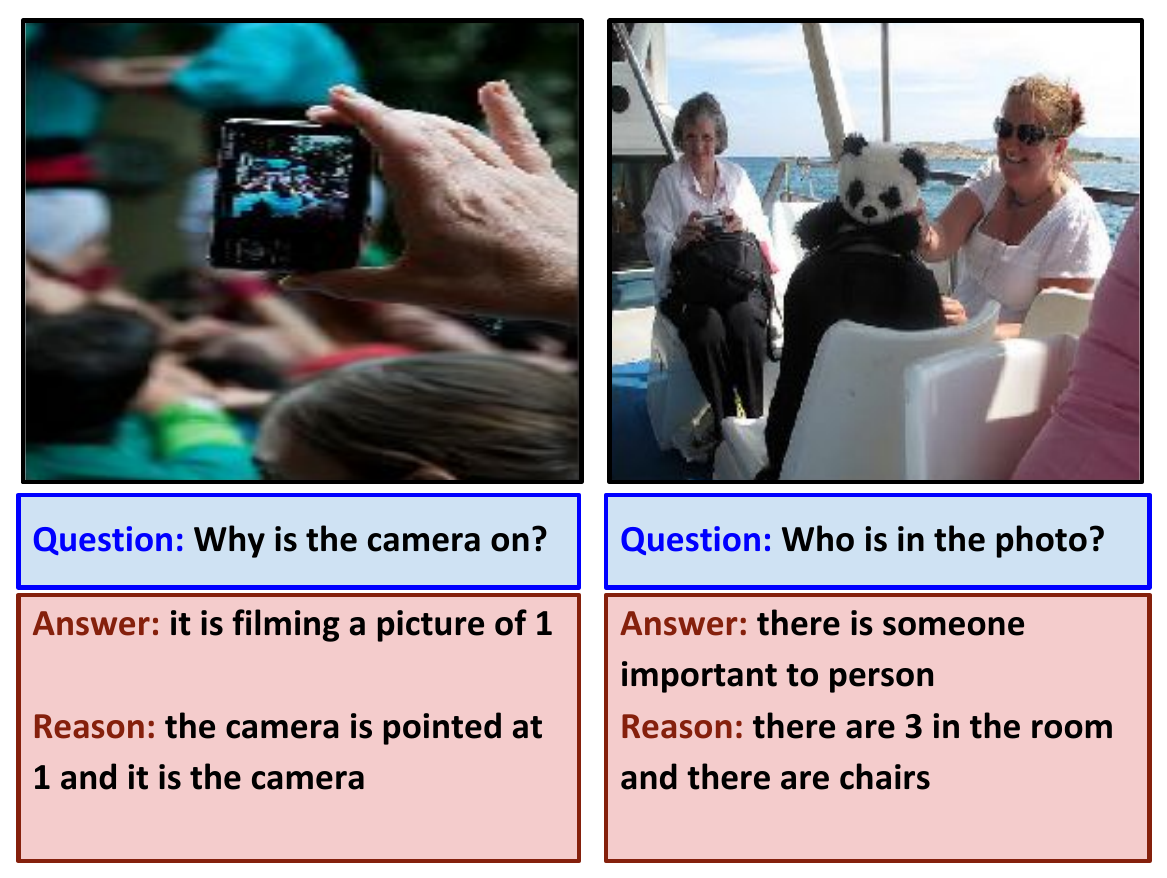}\label{fig:q1}}
\subfloat{\includegraphics[width=0.45\linewidth]{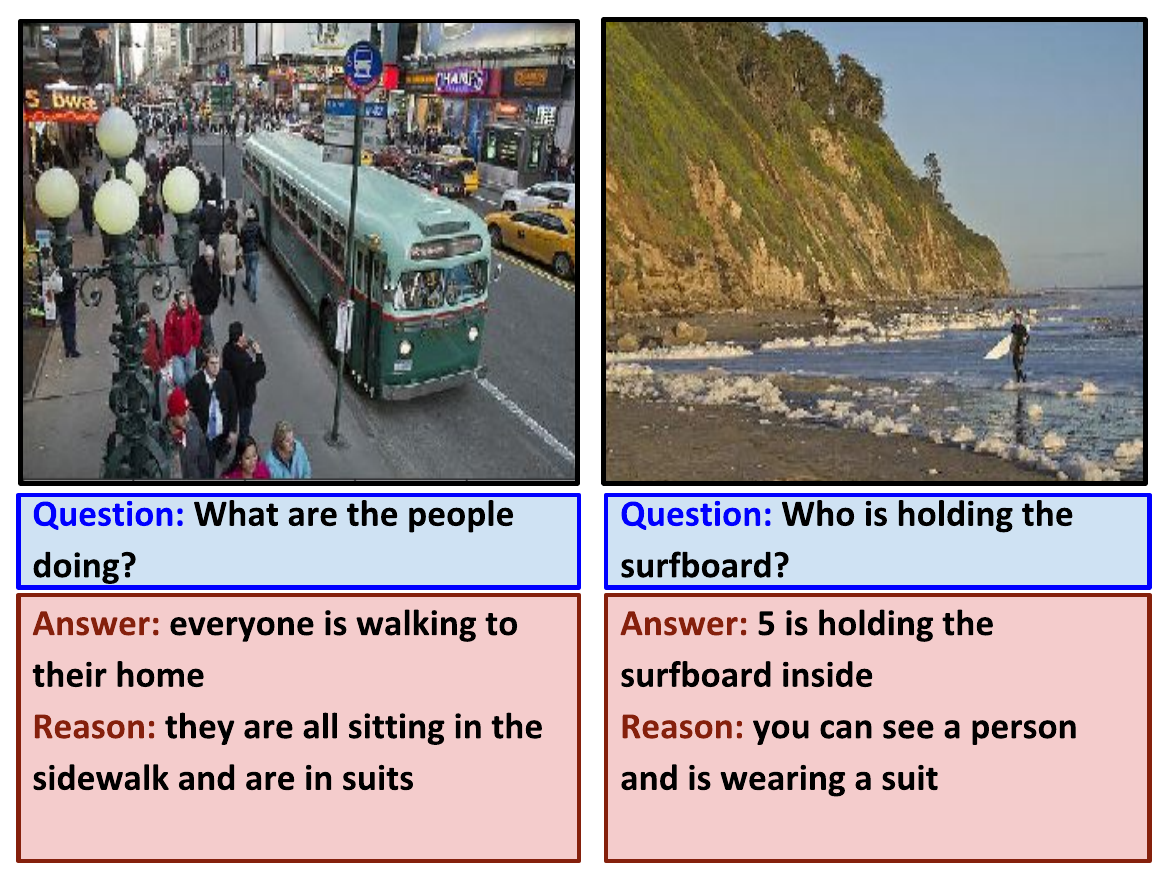}\label{fig:q2}}
% %\vspace{-6pt}
\caption{\textit{(Best viewed in color)} Qualitative results on Visual7W dataset for \pname\ task from our proposed Generation-Refinement architecture. Blue box = Question about image; Red box = Generated results from our proposed architecture. (Note that there is Reason provided in the Visual7W dataset, and all the reasons in the figures are generated by our model.)}
%\vspace{-6pt}
\label{fig:visual_7w_qual_results}
\end{figure*}

%\vspace{-4pt}
As stated in Section 6 of the main paper, we also studied whether the proposed model, trained on the VCR dataset, can provide answers and rationales to visual questions in standard VQA datasets (which do not have ground truth rationale provided). %To this end, we tested our trained model on the Visual7W [53] dataset without any additional training. 
% The Visual7W dataset provides ground truth answers to the questions but no rationale, hence limiting us from performing quantitative evaluation. Our qualitative results are however presented in
Figure \ref{fig:visual_7w_qual_results} presents additional qualitative results for \pname\ task on the Visual7W dataset. We observe that our algorithm generalizes reasonably well to the other VQA dataset and generates answers and rationales relevant to the image-question pair, without any explicit training for this dataset. This adds a promising dimension to this work.
\section{On objective evaluation metrics for generative tasks: A Discussion}
\begin{figure*}
\centering
\subfloat{\includegraphics[width=\linewidth]{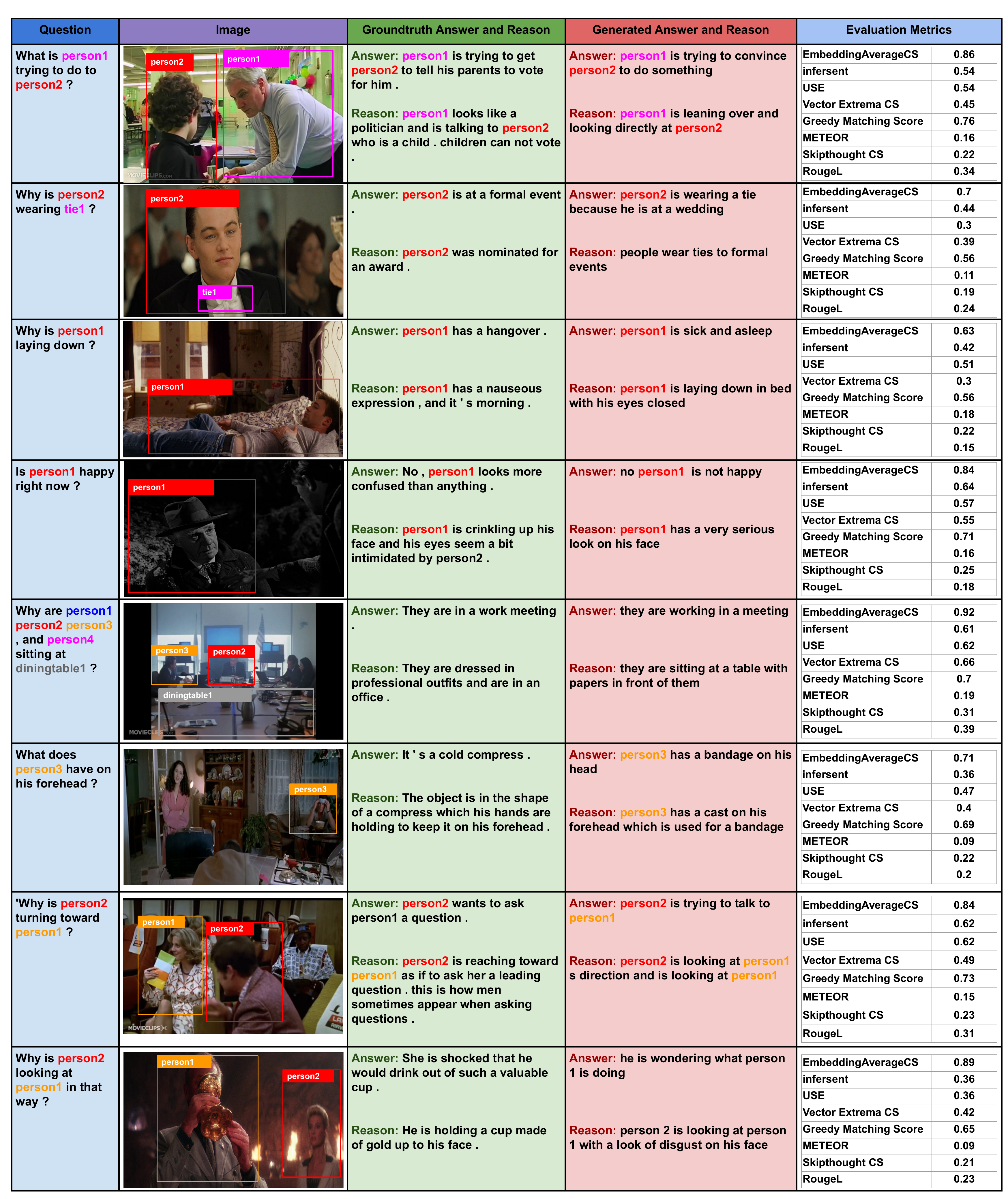}\label{fig:q1}}\hfill% %\vspace{-6pt}
\caption{\textit{(Best viewed in color)} Sample quantitative and qualitative results that show that evaluation metrics can have low scores even when results are qualitatively good. Blue box = question about image; Green = Ground truth; Red = Generated results from our proposed architecture.}
%\vspace{-6pt}
\label{fig:eval_metrics}
\end{figure*}

% Despite the remarkably good qualitative results, the values of evaluation metrics is not so high. In this section, we analyze in detail about the low quantitative results. The generated answer-rationale pair may be reasonably good, but they may be very different from the ground-truth answer-rationale pair. This is because for a question there may be many correct answer and rationale because the generated output depends on visual understanding and the related context. Therefore, we cannot completely rely on the automatic evaluation metrics (Described in Section 5)  which finds the similarity score between the generated and ground-truth sentences. Therefore we did a Turing Test (described in Section 5) which helped to determine that our proposed generation-refinement model generates good answer-rationale pair.
Since \pname\ is a completely generative task, objective evaluation is a challenge, as in any other generative methods. Hence, for comprehensive evaluation, we use a suite of several well-known objective evaluation metrics to measure the performance of our method quantitatively. There are various reasons why our approach may seem to give relatively lower scores than typical results for these scores on other language processing tasks. Such evaluation metrics measure the similarity between the generated and ground-truth sentences. For our task, there may be multiple correct answers and rationales, and each of them can be expressed in numerous ways. 
Figure~\ref{fig:eval_metrics} shows a few examples of images and questions along with their corresponding ground-truth, generated answer-rationale pair, and corresponding evaluation metric scores. We observe that generated answers and rationales are relevant to the image-question pair but may be different from the ground-truth answer-rationale pair. Hence, the evaluation metrics reported here have low scores even when the results are actually qualitatively good (as evidenced in the Human Turing test results in Section 5 of the main paper).  Thus, in this task, textual similarity to the ground truth may not be the only sign of the quality and may even indicate that the network is overfitting. We hence use Turing Tests (described in Section 5 of the main paper) to better estimate the performance of our model. An overall assessment that considers the different metrics used provides a more holistic view of the performance of our model.
% \redc{I have to add a table with few samples with their corresponding ground-truth, generated answer-rationale pair and corresponding evaluation metric scores to show that evaluation metric have low scores even when the results are qualitatively good.}
% \input{appendix/5_Model_Details.tex}
% \vspace{20pt}
% \input{appendix/7_Visual_atten_maps.tex}
% \bibliographystyle{named}
% \bibliography{references}
\end{document}